\batchmode
\makeatletter
\def\input@path{{"C:/Trabajo laptop/Mis articulos/Finished/Trajectory PMB filter/Accepted/"}}
\makeatother
\documentclass[british,english]{IEEEtran}
\usepackage[T1]{fontenc}
\usepackage[latin9]{inputenc}
\usepackage{array}
\usepackage{float}
\usepackage{multirow}
\usepackage{amsmath}
\usepackage{amsthm}
\usepackage{amssymb}
\usepackage{graphicx}

\makeatletter

\providecommand{\tabularnewline}{\\}
\floatstyle{ruled}
\newfloat{algorithm}{tbp}{loa}
\providecommand{\algorithmname}{Algorithm}
\floatname{algorithm}{\protect\algorithmname}

\theoremstyle{plain}
\newtheorem{thm}{\protect\theoremname}
\theoremstyle{definition}
\newtheorem{defn}[thm]{\protect\definitionname}
\theoremstyle{plain}
\newtheorem{prop}[thm]{\protect\propositionname}
\theoremstyle{plain}
\newtheorem{lem}[thm]{\protect\lemmaname}

\usepackage{cite}
\allowdisplaybreaks 

\usepackage[margin=8pt,font=footnotesize]{caption}
\usepackage{algorithm}
\usepackage{algpseudocode}
\usepackage{amsmath}  

\makeatother

\usepackage{babel}
\addto\captionsbritish{\renewcommand{\algorithmname}{Algorithm}}
\addto\captionsbritish{\renewcommand{\definitionname}{Definition}}
\addto\captionsbritish{\renewcommand{\lemmaname}{Lemma}}
\addto\captionsbritish{\renewcommand{\propositionname}{Proposition}}
\addto\captionsbritish{\renewcommand{\theoremname}{Theorem}}
\addto\captionsenglish{\renewcommand{\definitionname}{Definition}}
\addto\captionsenglish{\renewcommand{\lemmaname}{Lemma}}
\addto\captionsenglish{\renewcommand{\propositionname}{Proposition}}
\addto\captionsenglish{\renewcommand{\theoremname}{Theorem}}
\providecommand{\definitionname}{Definition}
\providecommand{\lemmaname}{Lemma}
\providecommand{\propositionname}{Proposition}
\providecommand{\theoremname}{Theorem}

\begin{document}
\title{Trajectory Poisson multi-Bernoulli filters}
\author{Ángel F. García-Fernández, Lennart Svensson, Jason L. Williams, Yuxuan
Xia, Karl Granström\thanks{A. F. García-Fernández is with the Department of Electrical Engineering and Electronics, University of Liverpool, Liverpool L69 3GJ, United Kingdom (angel.garcia-fernandez@liverpool.ac.uk). L. Svensson, Y. Xia, and K. Granström are with the Department of Electrical Engineering, Chalmers University of Technology, SE-412 96 Gothenburg, Sweden (firstname.lastname@chalmers.se). J. L. Williams is with the Commonwealth Scientific and Industrial Research Organization (jason.williams@data61.csiro.au).} }

\maketitle

\begin{abstract}
This paper presents two trajectory Poisson multi-Bernoulli (TPMB)
filters for multi-target tracking: one to estimate the set of alive
trajectories at each time step and another to estimate the set of
all trajectories, which includes alive and dead trajectories, at each
time step. The filters are based on propagating a Poisson multi-Bernoulli
(PMB) density on the corresponding set of trajectories through the
filtering recursion. After the update step, the posterior is a PMB
mixture (PMBM) so, in order to obtain a PMB density, a Kullback-Leibler
divergence minimisation on an augmented space is performed. The developed
filters are computationally lighter alternatives to the trajectory
PMBM filters, which provide the closed-form recursion for sets of
trajectories with Poisson birth model, and are shown to outperform
previous multi-target tracking algorithms. 
\end{abstract}


\section{Introduction}

Multitarget tracking (MTT) consists of inferring the trajectories
of an unknown number of targets that appear and disappear from a scene
of interest based on noisy sensor data \cite{Blackman_book99,Challa_book11}.
Multitarget tracking is a fundamental process of numerous applications
including advanced driver assistance systems, self-driving vehicles
\cite{Chen16}, air traffic monitoring \cite{Hurter19} and maritime
surveillance \cite{Meyer18}. There are many approaches to perform
multitarget tracking such as multiple hypothesis tracking \cite{Reid79,Brekke18},
joint probabilistic data association \cite{Bar-Shalom75} and the
random finite set (RFS) framework \cite{Mahler_book14}.

The traditional RFS approach to MTT is mainly concerned with multi-target
filtering, in which one aims to estimate the current set of targets,
without attempting to estimate target trajectories. In some scenarios,
targets may appear anywhere in the surveillance area, while in others,
targets may appear at localised areas, e.g., airports or doors. Both
types of scenarios can be handled by the appropriate choice of birth
model. The birth model also enables the corresponding filters to keep
information on potential targets that may have been occluded \cite[Fig. 6]{Granstrom20},
which is key information in certain applications such as self-driving
vehicles.

With Poisson point process (PPP) birth model, the solution to the
multi-target filtering problem is given by the Poisson multi-Bernoulli
mixture (PMBM) filter \cite{Williams15b,Angel18_b}. If the birth
model is multi-Bernoulli instead of Poisson, the filtering density
is given by the multi-Bernoulli mixture (MBM) filter, which corresponds
to the PMBM filtering recursion by setting the intensity of the Poisson
process to zero and adding Bernoulli components for newborn targets
in the prediction \cite{Angel18_b,Angel19_e}. An MBM can also be
written as a mixture in which Bernoulli components have deterministic
existence instead of probabilistic, giving rise to the MBM$_{01}$
filter \cite[Sec. IV]{Angel18_b}. Deterministic existence leads to
an exponential growth in the number of mixture components, which is
undesirable from a computational point of view. In general, a PMBM
is preferred over MBM/MBM$_{01}$ forms due to a more efficient representation
of the information on undetected targets, via the intensity of a Poisson
RFS, not limiting a priori the maximum number of new born targets
at each time step \cite{Angel19_e}, and being able to handle continuous-time
multi-target models \cite{Angel20}. 

Even though the PMBM filter provides a closed-form solution to the
multi-target filtering problem, it is also relevant to consider computationally
lighter filters such as the probability hypothesis density (PHD) filter,
cardinality PHD filters \cite{Mahler_book14}, and Poisson multi-Bernoulli
(PMB) filters \cite{Williams15b,Williams15}. Relations between the
PMB filter and the joint integrated data association filter \cite{Musicki04}
were given in \cite{Williams15b}.

Track building procedures for the above-mentioned unlabelled filters
can be obtained based on filter meta-data \cite{Panta09,Battistelli12,Williams15b},
i.e., information contained in the hypothesis trees. However, the
posterior itself only provides information about the current set of
targets, and not their trajectories. One approach to building trajectories
from posterior densities is to add unique labels to the target states
and form trajectories by linking target state estimates with the same
label \cite{Angel13,Vo13,Aoki16}. With labelled multi-Bernoulli birth,
the $\delta$-generalised labelled multi-Bernoulli ($\delta$-GLMB)
filter \cite{Vo13} provides the corresponding filtering density,
via a recursion that is similar to the MBM$_{01}$ filter recursion
\cite[Sec. IV]{Angel18_b}. A computationally lighter alternative
to the $\delta$-GLMB filter is the labelled multi-Bernoulli (LMB)
filter \cite{Reuter14}. Sequential track building approaches based
on labelling can work well in many cases but it is not always adequate
due to ambiguity in target-to-label associations, e.g., for independent
and identically (IID) cluster birth models \cite[Sec. II.B]{Angel20_b}\cite[Sec. III.B]{Lu17}.

The above track building problems can be solved by computing (multi-object)
densities on sets of trajectories \cite{Angel20_b}, rather than sets
of labelled targets. This approach has led to the following filters:
trajectory PMBM (TPMBM) filter \cite{Granstrom18,Granstrom19_prov2},
trajectory MBM (TMBM) filter \cite{Xia19_b}, trajectory MBM$_{01}$
(TMBM$_{01}$) filter \cite{Angel20_b}, and trajectory PHD (TPHD)
and CPHD (TCPHD) filters \cite{Angel19_f}. These filters are analogous
to their set of targets counterparts, but have the ability to estimate
trajectories from first principles, and the possibility of improving
the estimation of past states in the trajectories.  The trajectory-based
filters with multi-Bernoulli birth can be augmented to include labels,
without affecting the filtering recursion \cite[Sec. IV.A]{Angel20_b}.

\begin{figure}
\begin{centering}
\includegraphics[scale=0.6]{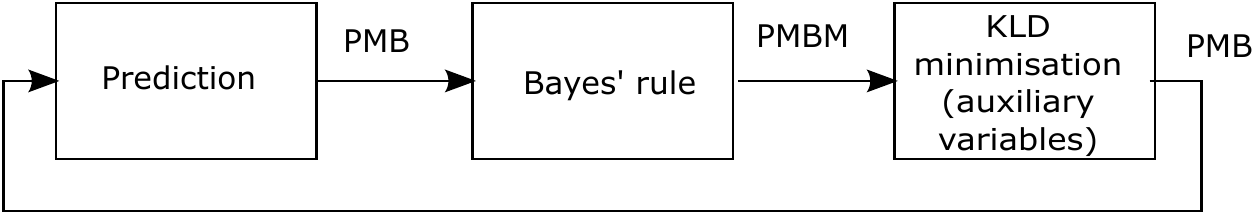}
\par\end{centering}
\caption{\label{fig:TPMB_diagram}Diagram of the two TPMB filters. The first
TPMB filter propagates a PMB density on the set of alive trajectories
at the current time. After each update step, the posterior is a PMBM
so the TPMB filter performs a KLD minimisation, in an augmented trajectory
space with an auxiliary variable. The second TPMB filter works equivalently
but propagating a PMB density on the set of all trajectories.}
\end{figure}

This paper proposes two trajectory PMB (TPMB) filters that approximate
the trajectory PMBM filters \cite{Granstrom18} using track-oriented
MBM merging \cite[Sec. IV.A]{Williams15b}. One TPMB filter aims to
estimate the set of the alive trajectories at each time step, while
the other aims to estimate the set of all trajectories (alive and
already dead) at each time step. Keeping probabilistic information
on all trajectories is important in many applications, for example,
surveillance and retail analytics \cite{Newman02}. In the TPMB filters,
the Poisson component represents information regarding trajectories
that have not been detected and the multi-Bernoulli component represents
information on trajectories that have been detected at some point
in the past. As the true posterior is a TPMBM, the TPMB filter is
derived by making use of a Kullback-Leibler divergence (KLD) minimisation,
on a trajectory space with auxiliary variables, after each update,
see Figure \ref{fig:TPMB_diagram}. The resulting TPMB density also
matches the PHD of the updated TPMBM. As the TPMB filtering posterior
is defined over the set of trajectories, one can estimate the set
of trajectories directly from this density. In this paper, we also
propose a Gaussian implementation of the TPMB filters for linear/Gaussian
models. Simulation results show that the TPMB filters have a performance
close to the TPMBM filters, with a decrease in computational complexity,
and outperform other filters in the literature.

The rest of the paper is organised as follows. We formulate the considered
multitarget tracking problems in Section \ref{sec:Problem-formulation}.
The proposed TPMB approximation to a TPMBM density using KLD minimisation
is obtained in Section \ref{sec:Best-trajectory-PMB}. The resulting
TPMB filters are proposed in Section \ref{sec:Trajectory-PMB-filters}
and their Gaussian implementations in Section \ref{sec:Gaussian-TPMB-filters}.
Simulation results are shown in Section \ref{sec:Simulations} and
conclusions are drawn in Section \ref{sec:Conclusions}.

\section{Problem formulation\label{sec:Problem-formulation}}

We tackle two multi-target tracking problems \cite{Granstrom18}: 
\begin{enumerate}
\item The estimation of the set of alive trajectories at the current time
step.
\item The estimation of the set of all trajectories that have existed up
to the current time step. We refer to this set as the set of all trajectories.
\end{enumerate}
These problems can be solved by calculating the (multi-trajectory)
density over the considered set of trajectories. In this paper, we
consider a computationally appealing approximation based on Poisson
multi-Bernoulli densities. 

\subsection{Set of trajectories\label{subsec:Set_trajectories}}

A single target state $x\in\mathbb{R}^{n_{x}}$ contains information
of interest about the target, e.g., its position and velocity. A set
of single target states $\mathbf{x}$ belongs to $\mathcal{F}\left(\mathbb{R}^{n_{x}}\right)$
where $\mathcal{F}\left(\mathbb{R}^{n_{x}}\right)$ denotes the set
of all finite subsets of $\mathbb{R}^{n_{x}}$. We are interested
in estimating target trajectories, where a trajectory consists of
a finite sequence of target states that can start at any time step
and end any time later on. A trajectory is therefore represented as
a variable $X=\left(t,x^{1:\nu}\right)$ where $t$ is the initial
time step of the trajectory, $\nu$ is its length and $x^{1:\nu}=\left(x^{1},...,x^{\nu}\right)$
denotes a finite sequence of length $\nu$ that contains the target
states.

We consider trajectories up to the current time step $k$. As a trajectory
$\left(t,x^{1:\nu}\right)$ exists from time step $t$ to $t+\nu-1$,
the variable $\left(t,\nu\right)$ belongs to the set $I_{(k)}=\left\{ \left(t,\nu\right):0\leq t\leq k\,\mathrm{and}\,1\leq\nu\leq k-t+1\right\} $.
A single trajectory $X$ up to time step $k$ therefore belongs to
the space $T_{\left(k\right)}=\uplus_{\left(t,\nu\right)\in I_{(k)}}\left\{ t\right\} \times\mathbb{R}^{\nu n_{x}}$,
where $\uplus$ stands for union of sets that are mutually disjoint.
We denote a set of trajectories up to time step $k$ as $\mathbf{X}\in\mathcal{F}\left(T_{\left(k\right)}\right)$.
Note that there can be multiple trajectories in $\mathbf{X}$ with
the same $\left(t,\nu\right)$. 

\subsubsection{Integrals and densities}

Given a real-valued function $\pi\left(\cdot\right)$ on the single
trajectory space $T_{\left(k\right)}$, its integral is \cite{Angel20_b}
\begin{align}
\int\pi\left(X\right)dX & =\sum_{\left(t,\nu\right)\in I_{(k)}}\int\pi\left(t,x^{1:\nu}\right)dx^{1:\nu}.\label{eq:single_trajectory_integral}
\end{align}
This integral goes through all possible start times, lengths and target
states of the trajectory. Given a real-valued function $\pi\left(\cdot\right)$
on the space $\mathcal{F}\left(T_{\left(k\right)}\right)$ of sets
of trajectories, its set integral is
\begin{align}
\int\pi\left(\mathbf{X}\right)\delta\mathbf{X} & =\sum_{n=0}^{\infty}\frac{1}{n!}\int\pi\left(\left\{ X_{1},...,X_{n}\right\} \right)dX_{1:n}\label{eq:set_integral_trajectory}
\end{align}
where $X_{1:n}=\left(X_{1},...,X_{n}\right)$. A function $\pi\left(\cdot\right)$
is a multitrajectory density of a random finite set of trajectories
if $\pi\left(\cdot\right)\geq0$ and its set integral is one. 

\subsection{Multi-target Bayesian models\label{subsec:Multi-target-Bayesian-models}}

The multi-target state evolves according to a Markov system with the
following modelling:
\begin{itemize}
\item P1 Given the set ${\bf x}_{k}$ of targets at time step $k$, each
target $x\in{\bf x}_{k}$ survives with probability $p^{S}\left(x\right)$
and moves to a new state with a transition density $g\left(\cdot\left|x\right.\right)$,
or dies with probability $1-p^{S}\left(x\right)$.
\item P2 The multitarget state ${\bf x}_{k+1}$ is the union of the surviving
targets and new targets, where new targets are born independently
following a PPP with intensity $\lambda^{B}\left(\cdot\right)$.
\end{itemize}
Set $\mathbf{x}_{k}$ is observed through the set $\mathbf{z}_{k}$
of measurements, which is modelled as
\begin{itemize}
\item U1 Each target state $x\in\mathbf{x}_{k}$ is either detected with
probability $p^{D}\left(x\right)$ and generates one measurement with
density $l\left(\cdot|x\right)$, or missed with probability $1-p^{D}\left(x\right)$. 
\item U2 The set $\mathbf{z}_{k}$ is the union of the target-generated
measurements and Poisson clutter with intensity $\lambda^{C}\left(\cdot\right)$. 
\end{itemize}
U1 and U2 imply that a measurement is generated by at most one target.
The objective is to compute the posterior density of the set $\mathbf{X}_{k}$
of trajectories at time step $k$ given the sequence $\left(\mathbf{z}_{1},...,\mathbf{z}_{k}\right)$
of measurements up to time step $k$. Depending on the problem formulation,
this set can correspond to the set of all trajectories or the set
of alive trajectories, and the meaning will be clear from context
so we use the same notation. If $\mathbf{X}_{k}$ denotes the set
of alive trajectories at time step $k$, then $t+\nu-1=k$ for each
$\left(t,x^{1:\nu}\right)\in\mathbf{X}_{k}$. If $\mathbf{X}_{k}$
denotes the set of all trajectories at time step $k$, then $t+\nu-1\leq k$
for each $\left(t,x^{1:\nu}\right)\in\mathbf{X}_{k}$.

\subsection{Poisson multi-Bernoulli mixture}

Given the sequence of measurements $\left(\mathbf{z}_{1},...,\mathbf{z}_{k}\right)$
up to time step $k$ and the models in Section \ref{subsec:Multi-target-Bayesian-models},
the density $f_{k'|k}\left(\cdot\right)$ of the set of trajectories
at time step $k'\in\left\{ k,k+1\right\} $ is a PMBM \cite{Granstrom18}.
This holds for the two problem formulations we consider, though the
specific recursions to compute $f_{k'|k}\left(\cdot\right)$ vary.
In both cases, $f_{k'|k}\left(\cdot\right)$ is a PMBM with
\begin{align}
f_{k'|k}\left(\mathbf{X}_{k'}\right) & =\sum_{\mathbf{Y}\uplus\mathbf{W}=\mathbf{X}_{k'}}f_{k'|k}^{p}\left(\mathbf{Y}\right)f_{k'|k}^{mbm}\left(\mathbf{W}\right)\label{eq:TPMBM_original_mult}\\
f_{k'|k}^{p}\left(\mathbf{X}_{k'}\right) & =e^{-\int\lambda_{k'|k}\left(X\right)dX}\left[\lambda_{k'|k}\left(\cdot\right)\right]^{\mathbf{X}_{k'}}\\
f_{k'|k}^{mbm}\left(\mathbf{X}_{k'}\right) & =\sum_{a\in\mathcal{A}_{k'|k}}w_{k'|k}^{a}\sum_{\uplus_{l=1}^{n_{k'|k}}\mathbf{X}^{l}=\mathbf{X}_{k'}}\prod_{i=1}^{n_{k'|k}}f_{k'|k}^{i,a^{i}}\left(\mathbf{X}^{i}\right)
\end{align}
where the sum in (\ref{eq:TPMBM_original_mult}) goes through all
disjoint and possibly empty sets $\mathbf{Y}$ and $\mathbf{W}$ such
that $\mathbf{Y}\cup\mathbf{W}=\mathbf{X}_{k'}$, the multi-object
exponential is defined as $h^{\mathbf{X}}=\prod_{X\in\mathbf{X}}h\left(X\right)$
where $h$ is a real-valued function and $h^{\emptyset}=1$ by convention,
and
\begin{align}
f_{k'|k}^{i,a^{i}}\left(\mathbf{X}\right) & =\begin{cases}
1-r_{k'|k}^{i,a^{i}} & \mathbf{X}=\emptyset\\
r_{k'|k}^{i,a^{i}}p_{k'|k}^{i,a^{i}}\left(X\right) & \mathbf{X}=\left\{ X\right\} \\
0 & \mathrm{otherwise}.
\end{cases}\label{eq:Bernoulli_density_filter-1}
\end{align}
We proceed to explain the aspects of (\ref{eq:TPMBM_original_mult})
that are relevant to our contribution; for details, we refer the reader
to \cite{Granstrom18,Williams15b}. 

From (\ref{eq:TPMBM_original_mult}), the PMBM is the union of two
independent RFS: a Poisson RFS with (multi-trajectory) density $f_{k'|k}^{p}\left(\cdot\right)$
that represents undetected trajectories, and a mixture of multi-Bernoulli
RFS $f_{k'|k}^{mbm}\left(\cdot\right)$ that represents trajectories
that have been detected at some point up to time step $k$. The Poisson
RFS on undetected targets/trajectories provides very valuable information
in some applications. For example, in self-driving vehicles the Poisson
RFS can indicate areas where there may be pedestrians or vehicles
that have not been yet detected by our sensors \cite{Granstrom20}.
The intensity of the Poisson RFS is denoted as $\lambda_{k'|k}\left(\cdot\right)$.
Each received measurement generates a unique Bernoulli component.
The number of Bernoulli components is $n_{k'|k}$, and they are indexed
by variable $i$. A global hypothesis is $a=\left(a^{1},...,a^{n_{k'|k}}\right)$,
where $a^{i}\in\left\{ 1,...,h^{i}\right\} $ is the index to the
local hypothesis for the $i$-th Bernoulli and $h^{i}$ is the number
of local hypotheses. The weight of global hypothesis $a$ is
\begin{align}
w_{k'|k}^{a} & =\frac{\prod_{i=1}^{n_{k'|k}}w_{k'|k}^{i,a^{i}}}{\sum_{b\in\mathcal{A}_{k'|k}}\prod_{i=1}^{n_{k'|k}}w_{k'|k}^{i,b^{i}}}\label{eq:weight_global}
\end{align}
where $w_{k'|k}^{i,a^{i}}$ is the weight of local hypothesis $a_{i}$
for Bernoulli $i$. The set $\mathcal{A}_{k'|k}$ contains all global
hypotheses \cite{Williams15b}. The density of the $i$-th Bernoulli
with local hypothesis $a^{i}$ is $f_{k'|k}^{i,a^{i}}\left(\cdot\right)$,
with probability $r_{k'|k}^{i,a^{i}}$ of existence and single-trajectory
density $p_{k'|k}^{i,a^{i}}\left(\cdot\right)$. We can also write
(\ref{eq:TPMBM_original_mult}) as
\begin{align}
 & f_{k'|k}\left(\mathbf{X}_{k'}\right)\nonumber \\
 & =\sum_{\uplus_{l=1}^{n_{k'|k}}\mathbf{X}^{l}\uplus\mathbf{Y}=\mathbf{X}_{k'}}f_{k'|k}^{p}\left(\mathbf{Y}\right)\sum_{a\in\mathcal{A}_{k'|k}}w_{k'|k}^{a}\prod_{i=1}^{n_{k'|k}}\left[f_{k'|k}^{i,a^{i}}\left(\mathbf{X}^{i}\right)\right].\label{eq:TPMBM_original}
\end{align}
We will use this formulation in Section \ref{subsec:PMBM-with-auxiliary},
where we introduce auxiliary variables.

\section{Trajectory PMB approximation\label{sec:Best-trajectory-PMB}}

In this section we first introduce auxiliary variables in the PMBM
(\ref{eq:TPMBM_original}) in Section \ref{subsec:PMBM-with-auxiliary}.
Then, we provide the best PMB approximation to the PMBM with auxiliary
variables, in the sense of minimising the resulting KLD in Section
\ref{subsec:KLD-minimisation}. Finally, in Section \ref{subsec:Relation-between-KLDs},
we show the relation between the KLDs with and without auxiliary variables.

\subsection{PMBM with auxiliary variables\label{subsec:PMBM-with-auxiliary}}

In this section, we introduce an auxiliary variable in the PMBM representation
in (\ref{eq:TPMBM_original}) that will be useful to obtain the PMB
approximation. Auxiliary variables are commonly used in Bayesian inference
to deal with mixtures of densities, especially in sampling-based methods
\cite{Pitt99,Elvira19,Angel13,Angel16,Ubeda17,Morelande07,Yi13,Uney18}
and latent variable models, e.g., expectation-maximisation \cite{Bishop_book06}.
Given (\ref{eq:TPMBM_original}), we extend the single trajectory
space with an auxiliary variable $u\in\mathbb{\mathbb{U}}_{k'|k}=\left\{ 0,1,..,n_{k'|k}\right\} $,
such that $\left(u,X\right)\in\mathbb{\mathbb{U}}_{k'|k}\times T_{\left(k'\right)}$.
As will become clearer at the end of this section, variable $u=0$
implies that the single trajectory has not yet been detected, so it
corresponds to the PPP, and $u=i$ indicates that the single trajectory
corresponds to the $i$-th Bernoulli component. We denote a set of
trajectories with auxiliary variables as $\widetilde{\mathbf{X}}_{k'}\in\mathcal{F}\left(\mathbb{\mathbb{U}}_{k'|k}\times T_{\left(k'\right)}\right)$. 
\begin{defn}
Given $f_{k'|k}\left(\cdot\right)$ of the form (\ref{eq:TPMBM_original}),
we define the density $\widetilde{f}_{k'|k}\left(\cdot\right)$ on
the space $\mathcal{F}\left(\mathbb{\mathbb{U}}_{k'|k}\times T_{\left(k'\right)}\right)$
of sets of trajectories with auxiliary variable as 
\begin{align}
 & \widetilde{f}_{k'|k}\left(\widetilde{\mathbf{X}}_{k'}\right)\nonumber \\
 & =\sum_{\uplus_{l=1}^{n_{k'|k}}\mathbf{\widetilde{X}}^{l}\uplus\widetilde{\mathbf{Y}}=\widetilde{\mathbf{X}}_{k'}}\widetilde{f}_{k'|k}^{p}\left(\widetilde{\mathbf{Y}}\right)\sum_{a\in\mathcal{A}_{k'|k}}w_{k'|k}^{a}\prod_{i=1}^{n_{k'|k}}\left[\widetilde{f}_{k'|k}^{i,a^{i}}\left(\mathbf{\widetilde{X}}^{i}\right)\right]\nonumber \\
 & =\widetilde{f}_{k'|k}^{p}\left(\widetilde{\mathbf{Y}}_{k'}\right)\sum_{a\in\mathcal{A}_{k'|k}}w_{k'|k}^{a}\prod_{i=1}^{n_{k'|k}}\left[\widetilde{f}_{k'|k}^{i,a^{i}}\left(\widetilde{\mathbf{X}}_{k'}^{i}\right)\right]\label{eq:PMBM_aux_var2}
\end{align}
where, for a given $\widetilde{\mathbf{X}}_{k'}$, $\widetilde{\mathbf{Y}}_{k'}=\left\{ \left(u,X\right)\in\widetilde{\mathbf{X}}_{k'}:u=0\right\} $
and $\widetilde{\mathbf{X}}_{k'}^{i}=\left\{ \left(u,X\right)\in\widetilde{\mathbf{X}}_{k'}:u=i\right\} $,
and
\begin{align}
\widetilde{f}_{k'|k}^{p}\left(\widetilde{\mathbf{X}}\right) & =e^{-\int\lambda_{k'|k}\left(x\right)dx}\left[\widetilde{\lambda}_{k'|k}\left(\cdot\right)\right]^{\widetilde{\mathbf{X}}}\label{eq:PPP_augmented}\\
\widetilde{\lambda}_{k'|k}\left(u,X\right) & =\delta_{0}\left[u\right]\lambda_{k'|k}\left(X\right)\\
\widetilde{f}_{k'|k}^{i,a^{i}}\left(\widetilde{\mathbf{X}}\right) & =\begin{cases}
1-r_{k'|k}^{i,a^{i}} & \widetilde{\mathbf{X}}=\emptyset\\
r_{k'|k}^{i,a^{i}}p_{k'|k}^{i,a^{i}}\left(X\right)\delta_{i}\left[u\right] & \widetilde{\mathbf{X}}=\left\{ \left(u,X\right)\right\} \\
0 & \mathrm{otherwise}\quad
\end{cases}\label{eq:Bernoulli_density_filter-aux}
\end{align}
where the Kronecker delta $\delta_{i}\left[u\right]=1$ if $u=i$
and $\delta_{i}\left[u\right]=0$, otherwise. Note that the sum over
sets disappears in (\ref{eq:PMBM_aux_var2}) as there is only one
possible partition of $\widetilde{\mathbf{X}}_{k'}$ into $\widetilde{\mathbf{Y}},$
$\mathbf{\widetilde{X}}^{1}$,..., $\mathbf{\widetilde{X}}^{n_{k'|k}}$
that provides a non-zero density. In addition, if two or more trajectories
in $\widetilde{\mathbf{X}}_{k'}$ have non-zero auxiliary variables
that are equal, then $\widetilde{f}_{k'|k}\left(\widetilde{\mathbf{X}}_{k'}\right)=0$,
as the corresponding Bernoulli component (\ref{eq:Bernoulli_density_filter-aux})
evaluated for more than one trajectory is zero. Conversely, there
can be multiple trajectories in $\widetilde{\mathbf{X}}_{k'}$ whose
auxiliary variable is zero, without implying that $\widetilde{f}_{k'|k}\left(\widetilde{\mathbf{X}}_{k'}\right)=0$.
Also, the density $\widetilde{f}_{k'|k}\left(\cdot\right)$ has domain
$\mathcal{F}\left(\mathbb{\mathbb{U}}_{k'|k}\times T_{\left(k'\right)}\right)$
which implies that we only consider auxiliary variables $u\in\mathbb{\mathbb{U}}_{k'|k}=\left\{ 0,1,..,n_{k'|k}\right\} $.

The definition of $\widetilde{f}_{k'|k}\left(\cdot\right)$ is mathematically
sound as it defines a density in $\mathcal{F}\left(\mathbb{\mathbb{U}}_{k'|k}\times T_{\left(k'\right)}\right)$
and, as we prove in App. \ref{sec:AppendixA}, integrating out the
auxiliary variables, we recover the original density. This procedure
of introducing auxiliary variables for PMBM for sets of trajectories
is also directly applicable for PMBM for sets of targets. In the target
case, the proposed use of auxiliary variables bears resemblance to
the approaches in \cite[Sec. IX]{Meyer18}\cite{Houssineau18,Cament20},
in which targets that have never been detected are indistinguishable,
in our case represented by the PPP with $u=0$, and targets that have
been detected are distinguishable, in our case represented by a unique
$u>0$ and a Bernoulli density. Therefore, $u$ can be considered
a mark \cite{Streit_book10}, but it is not a label, as in a labelled
approach \cite{Vo13}, each target must have a unique label. 
\end{defn}

\subsection{KLD minimisation with auxiliary variables\label{subsec:KLD-minimisation}}

In this section, we derive the best PMB fit of a PMBM using KLD minimisation
on the space with auxiliary variables. 

\subsubsection{Kullback-Leibler divergence}

Given a real-valued function $\pi\left(\cdot\right)$ on the single
trajectory space $\mathbb{\mathbb{U}}_{k'|k}\times T_{\left(k'\right)}$
with auxiliary variable, its single-trajectory integral is
\begin{align}
\int\pi\left(\widetilde{X}\right)d\widetilde{X} & =\sum_{u\in\mathbb{\mathbb{U}}_{k'|k}}\sum_{\left(t,\nu\right)\in I_{(k')}}\int\pi\left(u,t,x^{1:\nu}\right)dx^{1:\nu}.\label{eq:single_trajectory_integral_aux}
\end{align}
Given two densities $\widetilde{f}\left(\cdot\right)$ and $\widetilde{q}\left(\cdot\right)$
on the space $\mathcal{F}\left(\mathbb{\mathbb{U}}_{k'|k}\times T_{\left(k'\right)}\right)$
of sets of trajectories with auxiliary variable, the KLD $\mathrm{D}\left(\widetilde{f}\left\Vert \widetilde{q}\right.\right)$
is
\begin{align}
\mathrm{D}\left(\widetilde{f}\left\Vert \widetilde{q}\right.\right) & =\int\widetilde{f}\left(\widetilde{\mathbf{X}}\right)\log\frac{\widetilde{f}\left(\widetilde{\mathbf{X}}\right)}{\widetilde{q}\left(\widetilde{\mathbf{X}}\right)}\delta\widetilde{\mathbf{X}}.\label{eq:KLD}
\end{align}

\subsubsection{PMB approximation}

We aim to obtain a PMB approximation $\widetilde{q}\left(\cdot\right)$
on the space $\mathcal{F}\left(\mathbb{\mathbb{U}}_{k'|k}\times T_{\left(k'\right)}\right)$
such that
\begin{align}
\widetilde{q}\left(\widetilde{\mathbf{X}}_{k'}\right) & =\widetilde{q}^{p}\left(\widetilde{\mathbf{Y}}_{k'}\right)\prod_{i=1}^{n_{k'|k}}\left[\widetilde{q}^{i,1}\left(\widetilde{\mathbf{X}}_{k'}^{i}\right)\right]\label{eq:TPMB_approx}
\end{align}
where $\widetilde{q}^{p}\left(\cdot\right)$ is of the form (\ref{eq:PPP_augmented})
with intensity $\lambda^{q}\left(\cdot\right)$ and $\widetilde{q}^{i,1}\left(\cdot\right)$
is of the form (\ref{eq:Bernoulli_density_filter-aux}) with existence
probability $r^{i}$ and single-trajectory density $p^{i}\left(\cdot\right)$.

\begin{prop}
\label{prop:PMB_KLD_minimisation}Given a PMBM density $\widetilde{f}_{k'|k}\left(\cdot\right)$
of the form (\ref{eq:PMBM_aux_var2}), the parameters of the PMB density
$\widetilde{q}\left(\cdot\right)$, see (\ref{eq:TPMB_approx}), that
minimises the KLD $\mathrm{D}\left(\widetilde{f}_{k'|k}\left\Vert \widetilde{q}\right.\right)$
are given by 
\begin{align}
\lambda^{q}\left(X\right) & =\lambda_{k'|k}\left(X\right)\\
r^{i} & =\sum_{a^{i}=1}^{h^{i}}\overline{w}_{k'|k}^{i,a^{i}}r_{k'|k}^{i,a^{i}}\label{eq:existence_Prop2}\\
p^{i}\left(X\right) & =\frac{\sum_{a^{i}=1}^{h^{i}}\overline{w}_{k'|k}^{i,a^{i}}r_{k'|k}^{i,a^{i}}p_{k'|k}^{i,a^{i}}\left(X\right)}{\sum_{a^{i}=1}^{h^{i}}\overline{w}_{k'|k}^{i,a^{i}}r_{k'|k}^{i,a^{i}}}\label{eq:p_i_Prop2}
\end{align}
where
\begin{align}
\overline{w}_{k'|k}^{i,a^{i}} & =\sum_{b\in\mathcal{A}_{k'|k}:b^{i}=a^{i}}w_{k'|k}^{b}.\label{eq:weight_simplified_Prop2}
\end{align}
\end{prop}
Proposition \ref{prop:PMB_KLD_minimisation} is proved in App. \ref{sec:AppendixB}.
Note that we can also write $r^{i}$ and $p^{i}\left(\cdot\right)$
as in (\ref{eq:r_i_pmb_appendix}) and (\ref{eq:p_i_pmb_appendix}).
Nevertheless, (\ref{eq:existence_Prop2}) and (\ref{eq:p_i_Prop2})
are more suitable for implementation than (\ref{eq:r_i_pmb_appendix})
and (\ref{eq:p_i_pmb_appendix}), as the sum for $p^{i}\left(\cdot\right)$
in (\ref{eq:p_i_Prop2}) has a single term for each $p_{k'|k}^{i,a^{i}}\left(\cdot\right)$.
In App. \ref{sec:AppendixB}, we also show that the PHD of $\widetilde{q}\left(\cdot\right)$
matches the PHD of $\widetilde{f}_{k'|k}\left(\cdot\right)$. 

\subsection{Relation between KLDs\label{subsec:Relation-between-KLDs}}

In this section, we establish that the KLD on the space of sets of
trajectories with auxiliary variables is an upper bound on the KLD
for sets of trajectories without auxiliary variables. 
\begin{lem}
\label{lem:KLD_bound}Let $f_{k'|k}\left(\cdot\right)$ and $\widetilde{f}{}_{k'|k}\left(\cdot\right)$
be the PMBMs in (\ref{eq:TPMBM_original}) and (\ref{eq:PMBM_aux_var2}).
Let $q\left(\cdot\right)$ denote a multi-trajectory density and $\widetilde{q}\left(\cdot\right)$
an extension of $q\left(\cdot\right)$ with auxiliary variables. Then,
\begin{align}
\mathrm{D}\left(f_{k'|k}\left\Vert q\right.\right) & \leq\mathrm{D}\left(\widetilde{f}{}_{k'|k}\left\Vert \widetilde{q}\right.\right).\label{eq:KLD_bound}
\end{align}
\end{lem}
The proof of Lemma \ref{lem:KLD_bound} is given in App. \ref{sec:AppendixB}.
The KLD $\mathrm{D}\left(f_{k'|k}\left\Vert q\right.\right)$ is the
one we are primarily interested in, as it does not include auxiliary
variables. Nevertheless, introducing auxiliary variables enables us
to minimise the resulting KLD in closed-form, which is an upper bound
on $\mathrm{D}\left(f_{k'|k}\left\Vert q\right.\right)$. 

\section{Trajectory PMB filters\label{sec:Trajectory-PMB-filters}}

In Section \ref{subsec:Bayesian-models-set_trajectories}, we explain
the dynamic and measurement models written for sets of alive trajectories
and sets of all trajectories. The filtering recursions of the TPMB
filters are provided in Section \ref{subsec:Filtering-recursion-TPMB}.

\subsection{Bayesian models for sets of trajectories\label{subsec:Bayesian-models-set_trajectories}}

We proceed to write the Bayesian dynamic/measurement models for the
two types of problem formulations \cite{Granstrom18}. These models
are required for the filtering recursions in Section \ref{subsec:Filtering-recursion-TPMB}.
In particular, we specify the intensity $\lambda_{k+1}^{B}\left(X\right)$
of new born trajectories, the single trajectory transition density
$g_{k+1}\left(\cdot\left|X\right.\right)$ and the probability $p^{S}\left(X\right)$
of survival as a function of a trajectory $X$. In this paper, $p^{S}\left(X\right)$
refers to the probability that a trajectory remains in the considered
set of trajectories (either all trajectories or alive trajectories)
and is different from $p^{S}\left(x\right)$ as a function on a target
state $x$. 

\subsubsection{Dynamic model for the set of alive trajectories\label{subsec:Dynamic-model_alive}}

The set of alive trajectories evolves according to this Markov system
\begin{itemize}
\item P1T Given the set ${\bf X}_{k}$ of alive trajectories at time step
$k$, each $X=\left(t,x^{1:\nu}\right)\in{\bf X}_{k}$, where $t+\nu-1=k$,
either survives with probability $p^{S}\left(X\right)=p^{S}\left(x^{\nu}\right)$
with a transition density 
\begin{align}
g_{k+1}\left(t_{y},y{}^{1:\nu_{y}}\left|X\right.\right) & =\delta_{t}\left[t_{y}\right]\delta_{\nu+1}\left[\nu_{y}\right]\delta_{x^{1:\nu}}\left(y^{1:\nu_{y}-1}\right)\nonumber \\
 & \,\times g\left(y^{\nu_{y}}\left|x^{\nu}\right.\right)
\end{align}
or dies with probability $1-p^{S}\left(X\right)$.
\item P2T The  set ${\bf X}_{k+1}$ is the union of the surviving trajectories
and new trajectories, which are born independently following a PPP
with intensity 
\begin{align}
\lambda_{k+1}^{B}\left(t,x^{1:\nu}\right) & =\delta_{k+1}\left[t\right]\delta_{1}\left[\nu\right]\lambda^{B}\left(x^{\nu}\right).
\end{align}
\end{itemize}

\subsubsection{Dynamic model for the set of all trajectories\label{subsec:Dynamic-model_all}}

The set of all trajectories evolves according to this Markov system
\begin{itemize}
\item P3T Given the set ${\bf X}_{k}$ of all trajectories at time step
$k$, each $X=\left(t,x^{1:\nu}\right)\in{\bf X}_{k}$, where $t+\nu-1\leq k$,
survives with probability 1, $p^{S}\left(X\right)=1$, with a transition
density
\begin{align*}
 & g_{k+1}\left(t_{y},y{}^{1:\nu_{y}}\left|X\right.\right)=\delta_{t}\left[t_{y}\right]\\
 & \times\begin{cases}
\delta_{\nu}\left[\nu_{y}\right]\delta_{x^{1:\nu}}\left(y^{1:\nu_{y}}\right) & \omega_{y}<k\\
\delta_{\nu}\left[\nu_{y}\right]\delta_{x^{1:\nu}}\left(y^{1:\nu_{y}}\right)\left(1-p^{S}\left(x^{\nu}\right)\right) & \omega_{y}=k\\
\delta_{\nu+1}\left[\nu_{y}\right]\delta_{x^{1:\nu}}\left(y^{1:\nu_{y}-1}\right)p^{S}\left(x^{\nu}\right)\\
\quad\times g\left(y^{\nu_{y}}\left|x^{\nu}\right.\right) & \omega_{y}=k+1
\end{cases}
\end{align*}
where $\omega_{y}=t_{y}+\nu_{y}-1$.
\item Same birth model as in P2T.
\end{itemize}
It should be noted that $g_{k+1}\left(\cdot\right)$ does not depend
on $k$ in P1T (alive trajectories) but it depends on $k$ in P3T
(all trajectories). Nevertheless, we write the dependence on $k$
in both settings to have a unified notation for both transition densities.

\subsubsection{Measurement model for sets of trajectories\label{subsec:Measurement-model-trajectories}}

The measurement model U1 and U2 can be written for sets of all trajectories
and sets of alive trajectories in the same manner: 
\begin{itemize}
\item U1T Each trajectory $\left(t,x^{1:\nu}\right)\in\mathbf{X}_{k}$,
where $\mathbf{X}_{k}$ is the set of (all or alive) trajectories,
is detected with probability 
\begin{align}
p_{k}^{D}\left(t,x^{1:\nu}\right) & =\begin{cases}
p^{D}\left(x^{\nu}\right) & t+\nu-1=k\\
0 & \mathrm{otherwise}
\end{cases}
\end{align}
and generates one measurement with density $l\left(\cdot|t,x^{1:\nu}\right)=l\left(\cdot|x^{\nu}\right)$
or misdetected with probability $1-p_{k}^{D}\left(t,x^{1:\nu}\right)$. 
\item Same clutter model as in U2.
\end{itemize}

\subsection{Filtering recursions\label{subsec:Filtering-recursion-TPMB}}

We present the prediction and update in Sections \ref{subsec:TPMB_Prediction}
and \ref{subsec:TPMB_Update}. Given two real-valued functions $a\left(\cdot\right)$
and $b\left(\cdot\right)$ on the single-trajectory space, we denote
\begin{align}
\left\langle a,b\right\rangle  & =\int a\left(X\right)b\left(X\right)dX.
\end{align}

\subsubsection{Prediction\label{subsec:TPMB_Prediction}}

We denote the PMB filtering/predicted densities over the set of trajectories
at time step $k$, with $k'\in\left\{ k,k+1\right\} $, as
\begin{align}
f_{k'|k}\left(\mathbf{X}_{k'}\right) & =\sum_{\uplus_{l=1}^{n_{k'|k}}\mathbf{X}^{l}\uplus\mathbf{Y}=\mathbf{X}_{k'}}f_{k'|k}^{p}\left(\mathbf{Y}\right)\prod_{i=1}^{n_{k'|k}}\left[f_{k'|k}^{i}\left(\mathbf{X}^{i}\right)\right]\label{eq:PMB_k_1_k_1}
\end{align}
where the intensity of the Poisson component is $\lambda_{k'|k}\left(\cdot\right)$,
$n_{k'|k}$ is the number of Bernoulli components and the probability
of existence and single target density of the $i$-th Bernoulli component
are $r_{k'|k}^{i}$ and $p_{k'|k}^{i}\left(\cdot\right)$. 
\begin{lem}[TPMB prediction]
\label{lem:TPMB_prediction}Given the PMB filtering density on the
set of trajectories at time step $k-1$ of the form (\ref{eq:PMB_k_1_k_1}),
the predicted density at time step $k$ is a PMB of the form (\ref{eq:PMB_k_1_k_1}),
with $n_{k|k-1}=n_{k-1|k-1}$ and 
\begin{align}
\lambda_{k|k-1}\left(X\right) & =\lambda_{k}^{B}\left(X\right)+\left\langle \lambda_{k-1|k-1},g_{k}\left(X|\cdot\right)p^{S}\left(\cdot\right)\right\rangle \\
r_{k|k-1}^{i} & =r_{k-1|k-1}^{i}\left\langle p_{k-1|k-1}^{i},p^{S}\right\rangle \\
p_{k|k-1}^{i}\left(X\right) & =\frac{\left\langle p_{k-1|k-1}^{i},g_{k}\left(X|\cdot\right)p^{S}\left(\cdot\right)\right\rangle }{\left\langle p_{k-1|k-1}^{i},p^{S}\right\rangle }
\end{align}
where $g_{k}\left(\cdot|\cdot\right)$, $p^{S}\left(\cdot\right)$
and $\lambda_{B,k}\left(\cdot\right)$ are chosen depending on the
problem formulation: for alive trajectories, see Section \ref{subsec:Dynamic-model_alive},
and for all trajectories, see Section \ref{subsec:Dynamic-model_all}. 
\end{lem}
Lemma \ref{lem:TPMB_prediction} is a particular case of the TPMBM
predictor \cite{Granstrom18}, as a PMB is a PMBM with only one mixture
component.

\subsubsection{Update\label{subsec:TPMB_Update}}

The update of the TPMB filter is obtained by first doing a Bayesian
update, which yields a PMBM distribution, followed by a KLD minimisation,
on the space with auxiliary variables, as was illustrated in Figure
\ref{fig:TPMB_diagram}.
\begin{lem}[TPMB update]
\label{lem:TPMBM_update}Given the PMB predicted density on the set
of trajectories at time step $k$ of the form (\ref{eq:PMB_k_1_k_1}),
and a measurement set $\mathbf{z}_{k}=\left\{ z_{k}^{1},...,z_{k}^{m_{k}}\right\} $,
the updated distribution is a PMBM of the form (\ref{eq:TPMBM_original_mult})
where $n_{k|k}=n_{k|k-1}+m_{k}$ and 
\begin{align}
\lambda_{k|k}\left(X\right) & =\left(1-p_{k}^{D}\left(X\right)\right)\lambda_{k|k-1}\left(X\right).
\end{align}
For each Bernoulli component in $f_{k|k-1}\left(\cdot\right)$, $i\in\left\{ 1,...,n_{k|k-1}\right\} $,
there are $h^{i}=m_{k}+1$ local hypotheses, corresponding to a misdetection
and an update with one of the measurements. The misdetection hypothesis
for Bernoulli component $i\in\left\{ 1,...,n_{k|k-1}\right\} $ is
given by $\mathcal{M}\left(i,1\right)=\emptyset$,
\begin{align}
w_{k|k}^{i,1} & =1-r_{k|k-1}^{i}\left\langle p_{k|k-1}^{i},p_{k}^{D}\right\rangle \\
r_{k|k}^{i,1} & =\frac{r_{k|k-1}^{i}\left\langle p_{k|k-1}^{i},1-p_{k}^{D}\right\rangle }{1-r_{k|k-1}^{i}\left\langle p_{k|k-1}^{i},p_{k}^{D}\right\rangle }\\
p_{k|k}^{i,1}\left(X\right) & =\frac{\left(1-p_{k}^{D}\left(X\right)\right)p_{k|k-1}^{i}\left(X\right)}{\left\langle p_{k|k-1}^{i},1-p_{k}^{D}\right\rangle }.
\end{align}
The hypothesis for Bernoulli component $i\in\left\{ 1,...,n_{k|k-1}\right\} $
and measurement $z_{k}^{j}$ is given by $\mathcal{M}\left(i,j\right)=\left\{ j\right\} $,
$r_{k|k}^{i,1+j}=1$,
\begin{align}
w_{k|k}^{i,1+j} & =r_{k|k-1}^{i}\left\langle p_{k|k-1}^{i},l\left(z_{k}^{j}|\cdot\right)p_{k}^{D}\left(\cdot\right)\right\rangle \\
p_{k|k}^{i,1+j}\left(X\right) & =\frac{l\left(z_{k}^{j}|X\right)p_{k}^{D}\left(X\right)p_{k|k-1}^{i}\left(X\right)}{\left\langle p_{k|k-1}^{i},l\left(z_{k}^{j}|\cdot\right)p_{k}^{D}\left(\cdot\right)\right\rangle }.
\end{align}
For a new Bernoulli component $i\in\left\{ n_{k|k-1}+j\right\} $,
$j\in\left\{ 1,...,m_{k}\right\} $, which is initiated by measurement
$z_{k}^{j}$, there are $h_{i}=2$ local hypotheses
\begin{equation}
\mathcal{M}\left(i,1\right)=\emptyset,\,w_{k|k}^{i,1}=1,\,r_{k|k}^{i,1}=0,\:\mathcal{M}\left(i,2\right)=\left\{ j\right\} \label{eq:first_hypothesis_new_Bernoulli}
\end{equation}
\begin{align}
w_{k|k}^{i,2} & =\lambda^{C}\left(z_{k}^{j}\right)+\left\langle \lambda_{k|k-1},l\left(z_{k}^{j}|\cdot\right)p_{k}^{D}\left(\cdot\right)\right\rangle \label{eq:weight_new_born}\\
r_{k|k}^{i,2} & =\frac{\left\langle \lambda_{k|k-1},l\left(z_{k}^{j}|\cdot\right)p_{k}^{D}\left(\cdot\right)\right\rangle }{\lambda^{C}\left(z_{k}^{j}\right)+\left\langle \lambda_{k|k-1},l\left(z_{k}^{j}|\cdot\right)p_{k}^{D}\left(\cdot\right)\right\rangle }\\
p_{k|k}^{i,2}\left(X\right) & =\frac{l\left(z_{k}^{j}|X\right)p_{k}^{D}\left(X\right)\lambda_{k|k-1}\left(X\right)}{\left\langle \lambda_{k|k-1},l\left(z_{k}^{j}|\cdot\right)p_{k}^{D}\left(\cdot\right)\right\rangle }.
\end{align}
The set $\mathcal{M}\left(i,j\right)$ indicates the measurement index
for Bernoulli component $i$ and local hypothesis $j$. The set of
global data association hypotheses is 
\begin{align}
\mathcal{A}_{k} & =\left\{ \vphantom{\bigcup_{i=1}^{n_{k|k}}}\left(a_{1},...,a_{n_{k|k}}\right):a_{i}\in\mathbb{N}_{h_{i}},\bigcup_{i=1}^{n_{k|k}}\mathcal{M}\left(i,a_{i}\right)=\mathbb{N}_{m_{k}},\right.\nonumber \\
 & \left.\quad\mathcal{M}\left(i,a_{i}\right)\cap\mathcal{M}\left(j,a_{j}\right)=\emptyset\,\forall i\neq j\vphantom{\bigcup_{i=1}^{n_{k|k}}}\right\} \label{eq:global_hypotheses_PMB}
\end{align}
where $\mathbb{N}_{m_{k}}=\left\{ 1,...,m_{k}\right\} $.
\end{lem}
Lemma \ref{lem:TPMBM_update} is the same for both problem formulations
and corresponds to the TPMBM update \cite{Granstrom18} when the predicted
density is a PMB. Finally, the projection of this PMBM density to
a PMB density is obtained by Proposition \ref{prop:PMB_KLD_minimisation}. 

The results in Lemma \ref{lem:TPMB_prediction} and \ref{lem:TPMBM_update}
are conceptually analogous to the PMBM recursion for targets \cite{Williams15b,Angel18_b}.
However, these lemmas operate on sets of trajectories, which requires
using single trajectory densities and integrals, and establishing
the corresponding $p^{S}\left(\cdot\right)$, $g_{k+1}\left(\cdot\left|X\right.\right)$
and $\lambda_{k+1}^{B}\left(\cdot\right)$ depending on the problem
formulation, see Section \ref{subsec:Bayesian-models-set_trajectories}.
Also, the set of global hypotheses (\ref{eq:global_hypotheses_PMB})
is different from the global hypotheses of the PMBM recursion on sets
of targets \cite{Williams15b,Angel18_b}, as it considers a PMB prior,
not a PMBM prior.

\section{Gaussian TPMB filters\label{sec:Gaussian-TPMB-filters}}

In this section, we explain the Gaussian implementation of the TPMB
filter for alive trajectories and all trajectories in Section \ref{subsec:Gaussian-alive}
and \ref{subsec:Gaussian-all}, respectively. Practical considerations
are provided in Section \ref{subsec:Practical-considerations}. Trajectory
estimation is explained in Section \ref{subsec:Estimation}. Finally,
a discussion is provided in Section \ref{subsec:Discussion}. 

We use the notation
\begin{align}
\mathcal{N}\left(t,x^{1:\nu};\tau,\overline{x},P\right) & =\begin{cases}
\mathcal{N}\left(x^{1:\nu};\overline{x},P\right) & t=\tau,\,\nu=\iota\\
0 & \mathrm{otherwise}
\end{cases}\label{eq:Trajectory_Gaussian}
\end{align}
where $\iota=\mathrm{dim}\left(\overline{x}\right)/n_{x}$. Equation
(\ref{eq:Trajectory_Gaussian}) represents a single trajectory Gaussian
density with start time $\tau$, duration $\iota$, mean $\overline{x}\in\mathbb{R}^{\iota n_{x}}$
and covariance matrix $P\in\mathbb{R}^{\iota n_{x}\times\iota n_{x}}$
evaluated at $\left(t,x^{1:\nu}\right)$. We use $\otimes$ to indicate
the Kronecker product and $0_{m,n}$ is the $m\times n$ zero matrix. 

We make the additional assumptions
\begin{itemize}
\item A1 The survival and detection probabilities are constants: $p^{S}\left(x\right)=p^{S}$
and $p^{D}\left(x\right)=p^{D}$, see P1 and U1.
\item A2 $g\left(\cdot\left|x\right.\right)=\mathcal{N}\left(\cdot;Fx,Q\right)$
and $l\left(\cdot|x\right)=\mathcal{N}\left(\cdot;Hx,R\right)$.
\item A3 The PHD of the birth density at time step $k$ is
\begin{align}
\lambda_{k}^{B}\left(X\right) & =\sum_{q=1}^{n_{k}^{b}}w_{k}^{b,q}\mathcal{N}\left(X;k,\overline{x}_{k}^{b,q},P_{k}^{b,q}\right)\label{eq:GMPHD-birth}
\end{align}
where $n_{k}^{b}\in\mathbb{N}$ is the number of components, $w_{k}^{b,q}$
is the weight of the $q$th component, $\overline{x}_{k}^{b,q}\in\mathbb{R}^{n_{x}}$
its mean and $P_{k}^{b,q}\in\mathbb{R}^{n_{x}\times n_{x}}$ its covariance
matrix. 
\end{itemize}

\subsection{Gaussian implementation for alive trajectories\label{subsec:Gaussian-alive}}

In the Gaussian implementation for alive trajectories, the single-trajectory
density of the $i$-th Bernoulli component, see (\ref{eq:PMB_k_1_k_1}),
is of the form
\begin{align}
p_{k'|k}^{i}\left(X\right) & =\mathcal{N}\left(X;t^{i},\overline{x}_{k'|k}^{i},P_{k'|k}^{i}\right)\label{eq:single_trajectory_Gaussian_alive}
\end{align}
where $t^{i}$ is the start time, $\overline{x}_{k'|k}^{i}$ is the
mean, $P_{k'|k}^{i}$ the covariance matrix and $t^{i}+\mathrm{dim}\left(\overline{x}_{k'|k}^{i}\right)/n_{x}-1=k'$,
which implies that the trajectory is alive at time step $k'$, 

The PPP has a Gaussian mixture intensity
\begin{align}
\lambda_{k'|k}\left(X\right) & =\sum_{q=1}^{n_{k'|k}^{p}}w_{k'|k}^{p,q}\mathcal{N}\left(X;t_{k'|k}^{p,q},\overline{x}_{k'|k}^{p,q},P_{k'|k}^{p,q}\right)\label{eq:intensity_Gaussian_alive}
\end{align}
where $n_{k'|k}^{p}$ is the number of components, $w_{k'|k}^{p,q}$
is the weight of the $q$th component, $t_{k'|k}^{p,q}$ is starting
time, $\overline{x}_{k'|k}^{p,q}$ its mean and $P_{k'|k}^{p,q}$
its covariance matrix.

The prediction step is given by the following lemma.
\begin{lem}[GTPMB prediction, alive trajectories]
\label{lem:Prediction_Gaussian_TPMB_alive}Assume the filtering density
for the alive trajectories is a PMB (\ref{eq:PMB_k_1_k_1}) with $p_{k-1|k-1}^{i}\left(\cdot\right)$
and $\lambda_{k-1|k-1}\left(\cdot\right)$ given by (\ref{eq:single_trajectory_Gaussian_alive})
and (\ref{eq:intensity_Gaussian_alive}). Then, the predicted density
is a PMB of the form (\ref{eq:PMB_k_1_k_1}) with
\begin{align}
r_{k|k-1}^{i} & =p^{S}r_{k-1|k-1}^{i}\\
\overline{x}_{k|k-1}^{i} & =\left[\left(\overline{x}_{k-1|k-1}^{i}\right)^{T},\left(\overline{F}_{i}\overline{x}_{k-1|k-1}^{i}\right)^{T}\right]^{T}\label{eq:mean_prediction}\\
P_{k|k-1}^{i} & =\left[\begin{array}{cc}
P_{k-1|k-1}^{i} & P_{k-1|k-1}^{i}\overline{F}_{i}^{T}\\
\overline{F}_{i}P_{k-1|k-1}^{i} & \overline{F}_{i}P_{k-1|k-1}^{i}\overline{F}_{i}^{T}+Q
\end{array}\right]\label{eq:cov_prediction}\\
\overline{F}_{i} & =\left[0_{1,\iota^{i}-1},1\right]\otimes F,\\
\lambda_{k|k-1}\left(X\right) & =\sum_{q=1}^{n_{k}^{b}}w_{k}^{b,q}\mathcal{N}\left(X;k,\overline{x}_{k}^{b,q},P_{k}^{b,q}\right)+p^{S}\sum_{q=1}^{n_{k-1|k-1}^{p}}\nonumber \\
 & \,w_{k-1|k-1}^{p,q}\mathcal{N}\left(X;t_{k-1|k-1}^{p,q},\overline{x}_{k|k-1}^{p,q},P_{k|k-1}^{p,q}\right)\label{eq:lambda_prediction}
\end{align}
where $\iota^{i}=\mathrm{dim}\left(\overline{x}_{k-1|k-1}^{i}\right)/n_{x}$
and $\overline{x}_{k|k-1}^{p,q}$ and $P_{k|k-1}^{p,q}$ are obtained
by (\ref{eq:mean_prediction}) and (\ref{eq:cov_prediction}) using
$\overline{x}_{k-1|k-1}^{p,q}$ and $P_{k-1|k-1}^{p,q}$ instead of
$\overline{x}_{k-1|k-1}^{i}$ and $P_{k-1|k-1}^{i}$.
\end{lem}
The update step is given by the following lemma. 
\begin{lem}[GTPMB update, alive trajectories]
\label{lem:Update_Gaussian_TPMB_alive}Assume the PMB predicted density
(\ref{eq:PMB_k_1_k_1}) with $p_{k|k-1}^{i}\left(\cdot\right)$ and
$\lambda_{k|k-1}\left(\cdot\right)$ given by (\ref{eq:single_trajectory_Gaussian_alive})
and (\ref{eq:intensity_Gaussian_alive}). The updated density is a
PMBM. The PHD of the Poisson component is
\begin{align}
\lambda_{k|k}\left(X\right) & =\left(1-p^{D}\right)\lambda_{k|k-1}\left(X\right).\label{eq:lambda_update}
\end{align}
The misdetection hypothesis for Bernoulli component $i$ has
\begin{equation}
w_{k|k}^{i,1}=1-r_{k|k-1}^{i}p^{D},\:r_{k|k}^{i,1}=\frac{r_{k|k-1}^{i}\left(1-p^{D}\right)}{1-r_{k|k-1}^{i}p^{D}}
\end{equation}
\begin{align}
p_{k|k}^{i,1}\left(X\right) & =\mathcal{N}\left(X;t^{i},\overline{u}_{k|k}^{i,1},W_{k|k}^{i,1}\right)
\end{align}
where $\overline{u}_{k|k}^{i,1}=\overline{x}_{k|k-1}^{i}$ and $W_{k|k}^{i,1}=P_{k|k-1}^{i}$.
The detection hypothesis for Bernoulli component $i$ and measurement
$z_{k}^{j}$ has $r_{k|k}^{i,1+j}=1$,
\begin{align}
w_{k|k}^{i,1+j} & =r_{k|k-1}^{i}p^{D}\mathcal{N}\left(z_{k}^{j};\overline{z}_{i},S_{i}\right)\\
p_{k|k}^{i,1+j}\left(X\right) & =\mathcal{N}\left(X;t^{i},\overline{u}_{k|k}^{i,j},W_{k|k}^{i,j}\right)
\end{align}
\begin{align}
\overline{z}_{i} & =\overline{H}_{i}\overline{x}_{k|k-1}^{i},\quad S_{i}=\overline{H}_{i}P_{k|k-1}^{i}\overline{H}_{i}^{T}+R\label{eq:update_detected_Gaussian_ini}\\
\overline{H}_{i} & =\left[0_{1,\iota^{i}-1},1\right]\otimes H\\
\overline{u}_{k|k}^{i,j} & =\overline{x}_{k|k-1}^{i}+P_{k|k-1}^{i}\overline{H}_{i}^{T}S_{i}^{-1}\left(z_{k}^{j}-\overline{z}_{i}\right)\\
W_{k|k}^{i,j} & =P_{k|k-1}^{i}-P_{k|k-1}^{i}\overline{H}_{i}^{T}S_{i}^{-1}\overline{H}_{i}P_{k|k-1}^{i}.\label{eq:update_detected_Gaussian_end}
\end{align}
and $\iota^{i}=\mathrm{dim}\left(\overline{x}_{k|k-1}^{i}\right)/n_{x}$. 

For the new Bernoulli component $i$ initiated by measurement $z_{k}^{j}$,
the first local hypothesis has the parameters in (\ref{eq:first_hypothesis_new_Bernoulli}).
For the second hypothesis, we first calculate $v^{q}=\left\langle \lambda_{k|k-1},l\left(z_{k}^{j}|\cdot\right)p_{k}^{D}\right\rangle $
in (\ref{eq:weight_new_born}) for each PHD component (\ref{eq:intensity_Gaussian_alive})
$q\in\left\{ 1,...,n_{k|k-1}^{p}\right\} $,
\begin{align}
v^{q} & =p^{D}\mathcal{N}\left(z_{k}^{j};\overline{H}_{q}\overline{x}_{k|k-1}^{p,q},S_{q}\right)\label{eq:update_new_Gaussian_ini}\\
\overline{H}_{q} & =\left[0_{1,\iota^{q}-1},1\right]\otimes H,\quad S_{q}=\overline{H}_{q}P_{k|k-1}^{p,q}\overline{H}_{q}^{T}+R.
\end{align}
Then, we obtain $q^{*}=\max_{q}\left(v^{q}\right)$ and set
\begin{equation}
w_{k|k}^{i,2}=\lambda^{C}\left(z_{k}^{j}\right)+\sum_{q=1}^{n_{k|k-1}^{p}}v^{q},\;r_{k|k}^{i,2}=\frac{\sum_{q=1}^{n_{k|k-1}^{p}}v^{q}}{w_{k|k}^{i,2}}
\end{equation}
\begin{align}
p_{k|k}^{i,2}\left(X\right) & =\mathcal{N}\left(X;t_{k|k-1}^{p,q^{*}},\overline{u}_{k|k}^{i,2},W_{k|k}^{i,2}\right)\label{eq:density_new_Gaussian_alive}
\end{align}
\begin{align}
\overline{u}_{k|k}^{i,2} & =\overline{x}_{k|k-1}^{p,q^{*}}+P_{k|k-1}^{p,q^{*}}\overline{H}_{q^{*}}^{T}S_{q^{*}}^{-1}\left(z_{k}^{j}-\overline{H}_{q^{*}}\overline{x}_{k|k-1}^{p,q^{*}}\right)\\
W_{k|k}^{i,2} & =P_{k|k-1}^{p,q^{*}}-P_{k|k-1}^{p,q^{*}}\overline{H}_{q^{*}}^{T}S_{q^{*}}^{-1}\overline{H}_{q^{*}}P_{k|k-1}^{p,q^{*}}.\label{eq:update_new_Gaussian_end}
\end{align}
\end{lem}
In Lemma \ref{lem:Update_Gaussian_TPMB_alive}, all equations are
closed-form expressions obtained from Lemma \ref{lem:TPMBM_update},
except the single trajectory density for the new Bernoulli components
(\ref{eq:density_new_Gaussian_alive}). The closed-form formula of
the single trajectory densities of new Bernoulli components is actually
a Gaussian mixture, with potentially different starting times and
lengths. The filter becomes computationally more efficient by making
a Gaussian approximation. To do so, we take the Gaussian component
with highest weight (whose index is $q^{*}$) to obtain (\ref{eq:density_new_Gaussian_alive}).
This procedure was referred to as absorption in \cite{Angel19_f}.

After the Bayesian update, the updated density is a PMBM so a PMB
density is obtained by applying Proposition \ref{prop:PMB_KLD_minimisation},
see Figure \ref{fig:TPMB_diagram}. The resulting PMB has the same
PPP intensity as the updated PMBM. The resulting single-trajectory
densities $p_{k|k}^{i}\left(\cdot\right)$ are Gaussian mixtures so
we perform another KLD minimisation to fit a Gaussian distribution,
which is achieved by moment matching \cite{Bishop_book06}. The resulting
mean and covariance matrix are provided in App. \ref{sec:AppendixC}.

\subsection{Gaussian implementation for all trajectories\label{subsec:Gaussian-all}}

In the Gaussian implementation for the set of all trajectories, we
consider information over all trajectories that have ever been detected
and information regarding alive trajectories that have not been detected.
That is, as in most practical cases, trajectories that have never
been detected and no longer exist are not of importance, the PPP only
considers alive trajectories. The intensity of the PPP has the form
(\ref{eq:intensity_Gaussian_alive}) and the $i$-th Bernoulli component
has a single-trajectory density
\begin{align}
p_{k'|k}^{i}\left(X\right) & =\sum_{l=t^{i}}^{k'}\beta_{k'|k}^{i}\left(l\right)\mathcal{N}\left(X;t^{i},\overline{x}_{k'|k}^{i}\left(l\right),P_{k'|k}^{i}\left(l\right)\right)\label{eq:single_trajectory_Gaussian_all}
\end{align}
where $t^{i}$ is the start time, $\beta_{k'|k}^{i}\left(l\right)$
is the probability that the corresponding trajectory terminates at
time step $l$ (conditioned on existence), and $\overline{x}_{k'|k}^{i}\left(l\right)\in\mathbb{R}^{\iota n_{x}}$
and $P_{k'|k}^{i}\left(l\right)\in\mathbb{R}^{\iota n_{x}\times\iota n_{x}}$,
with $\iota=l-t^{i}+1$, are the mean and the covariance matrix of
the trajectory given that it ends at time step $l$. It should be
noted that $\sum_{l=t^{i}}^{k'}\beta_{k'|k}^{i}\left(l\right)=1$.

The prediction step is given by the following lemma.
\begin{lem}[GTPMB prediction, all trajectories]
\label{lem:Prediction_Gaussian_TPMB_all}Assume the PMB filtering
density for all trajectories (\ref{eq:PMB_k_1_k_1}) with $p_{k-1|k-1}^{i}\left(\cdot\right)$
and $\lambda_{k-1|k-1}\left(\cdot\right)$ given by (\ref{eq:single_trajectory_Gaussian_all})
and (\ref{eq:intensity_Gaussian_alive}). Then, the predicted density
is a PMB (\ref{eq:PMB_k_1_k_1}) with $\lambda_{k|k-1}\left(\cdot\right)$
given by (\ref{eq:lambda_prediction}) and $r_{k|k-1}^{i}=r_{k-1|k-1}^{i}$.
The means and covariance matrices of (\ref{eq:single_trajectory_Gaussian_all}),
for $l\in\left\{ t^{i},...,k-1\right\} $, are $\overline{x}_{k|k-1}^{i}\left(l\right)=\overline{x}_{k-1|k-1}^{i}\left(l\right)$
and $P_{k|k-1}^{i}\left(l\right)=P_{k-1|k-1}^{i}\left(l\right)$,
and, for $l=k$, $\overline{x}_{k|k-1}^{i}\left(k\right)$ and $P_{k|k-1}^{i}\left(k\right)$
are obtained substituting $\overline{x}_{k|k-1}^{i}\left(k-1\right)$
and $P_{k|k-1}^{i}\left(k-1\right)$ into (\ref{eq:mean_prediction})
and (\ref{eq:cov_prediction}). Finally,
\begin{align}
\beta_{k|k-1}^{i}\left(l\right) & =\begin{cases}
\beta_{k-1|k-1}^{i}\left(l\right) & l\in\left\{ t^{i},...,k-2\right\} \\
\left(1-p^{S}\right)\beta_{k-1|k-1}^{i}\left(l\right) & l=k-1\\
p^{S}\beta_{k-1|k-1}^{i}\left(k-1\right) & l=k.
\end{cases}
\end{align}
\end{lem}
It is important to notice that the prediction step does not modify
the single trajectory density of hypotheses that consider dead trajectories
$l\leq k-2$. For the hypothesis that considers that the trajectory
dies, $l=k-1$, the single trajectory density remains unchanged but
there is a change in the probability $\beta_{k'|k}^{i}\left(l\right)$,
as one has to take into account the probability of death $1-p^{S}$
. For the alive hypothesis, $l=k$, the mean and covariance matrix
are propagated as in the case of alive trajectories, see Lemma \ref{lem:Prediction_Gaussian_TPMB_alive},
and its corresponding probability $\beta_{k|k-1}^{i}\left(l\right)$
is obtained using the probability of survival and the probability
$\beta_{k-1|k-1}^{i}\left(k-1\right)$ that the trajectory was alive
at the previous time step.

The update step is given by the following lemma. 
\begin{lem}[GTPMB update, all trajectories]
\label{lem:Update_Gaussian_TPMB_all}Assume the PMB predicted density
(\ref{eq:PMB_k_1_k_1}) with $p_{k|k-1}^{i}\left(\cdot\right)$ and
$\lambda_{k|k-1}\left(\cdot\right)$ given by (\ref{eq:single_trajectory_Gaussian_all})
and (\ref{eq:intensity_Gaussian_alive}). Then, the updated density
is a PMBM. The PPP intensity $\lambda_{k|k}\left(\cdot\right)$ is
given by (\ref{eq:lambda_update}). The misdetection hypothesis for
Bernoulli component $i$ has the following parameters. The mean and
covariance matrices for $l\in\left\{ t^{i},...,k\right\} $ are $\overline{u}_{k|k}^{i,1}\left(l\right)=\overline{x}_{k|k-1}^{i}\left(l\right)$
and $W_{k|k}^{i,1}\left(l\right)=P_{k|k-1}^{i}\left(l\right)$, and
\begin{align}
w_{k|k}^{i,1} & =1-r_{k|k-1}^{i}\beta_{k|k-1}^{i}\left(k\right)p^{D}\\
r_{k|k}^{i,1} & =\frac{r_{k|k-1}^{i}\left(1-\beta_{k|k-1}^{i}\left(k\right)p^{D}\right)}{1-r_{k|k-1}^{i}\beta_{k|k-1}^{i}\left(k\right)p^{D}}\\
\beta_{k|k}^{i,1}\left(l\right) & \propto\begin{cases}
\beta_{k|k-1}^{i}\left(l\right) & l\in\left\{ t^{i},...,k-1\right\} \\
\left(1-p^{D}\right)\beta_{k|k-1}^{i}\left(l\right) & l=k.
\end{cases}
\end{align}
The detection hypothesis for Bernoulli component $i$ and measurement
$z_{k}^{j}$ has $r_{k|k}^{i,1+j}=1$
\begin{align}
w_{k|k}^{i,1+j} & =r_{k|k-1}^{i}\beta_{k|k-1}^{i}\left(k\right)p^{D}\mathcal{N}\left(z_{k}^{j};\overline{z}_{i},S_{i}\right)\label{eq:detected_weight_all_trajectory}\\
\beta_{k|k}^{i,1+j}\left(l\right) & =\begin{cases}
0 & l\in\left\{ t^{i},...,k-1\right\} \\
1 & l=k
\end{cases}\\
p_{k|k}^{i,1+j}\left(X\right) & =\mathcal{N}\left(X;t^{i},\overline{u}_{k|k}^{i,j},W_{k|k}^{i,j}\right)
\end{align}
where $\overline{z}_{i}$, $S_{i}$, $\overline{u}_{k|k}^{i,j}\left(k\right)$
and $W_{k|k}^{i,j}\left(k\right)$ are given by substituting $\overline{x}_{k|k-1}^{i}\left(k\right)$
and $P_{k|k-1}^{i}\left(k\right)$ into (\ref{eq:update_detected_Gaussian_ini})-(\ref{eq:update_detected_Gaussian_end}).

As we only consider alive trajectories in the PPP, for the new Bernoulli
component $i\in\left\{ n_{k|k-1}+j\right\} $ initiated by measurement
$z_{k}^{j}$, the update is done as in Lemma \ref{lem:Update_Gaussian_TPMB_alive},
which uses (\ref{eq:update_new_Gaussian_ini})-(\ref{eq:update_new_Gaussian_end}),
and  setting $\beta_{k|k}^{i,2}\left(k\right)=1$ and $\beta_{k|k}^{i,2}\left(l\right)=0$
$\forall l\neq k$.
\end{lem}
As happened with sets of alive trajectories, the updated density is
a PMBM, so we fit a PMB by applying Proposition \ref{prop:PMB_KLD_minimisation},
which keeps the PPP unaltered. The existence probability of the $i$-th
Bernoulli component is given by (\ref{eq:existence_Prop2}). The parameter
$\beta_{k|k}^{i}\left(\cdot\right)$ in (\ref{eq:single_trajectory_Gaussian_all})
becomes
\begin{align}
\beta_{k|k}^{i}\left(l\right) & =\sum_{a^{i}=1:r_{k|k}^{i,a^{i}}>0}^{h^{i}}\left[\frac{\overline{w}_{k|k}^{i,a^{i}}r_{k|k}^{i,a^{i}}}{r_{k|k}^{i}}\beta_{k|k}^{i,a^{i}}\left(l\right)\right]\label{eq:beta_KLD_minimisation}
\end{align}
for $l\in\left\{ t^{i},...,k\right\} $ and $\beta_{k|k}^{i}\left(l\right)=0$
otherwise. For each Bernoulli component, the hypotheses that represent
dead trajectories $l\in\left\{ t^{i},...,k-1\right\} $ are the same
for all $a\in\mathcal{A}_{k}$, with mean $\overline{u}_{k|k}^{i,1}\left(l\right)$
and covariance matrix $W_{k|k}^{i}\left(l\right)$, so the output
of Proposition \ref{prop:PMB_KLD_minimisation} is already in Gaussian
form for $l\in\left\{ t^{i},...,k-1\right\} $. For alive trajectories,
$l=k$, we perform moment matching to obtain the updated mean $\overline{x}_{k|k}^{i}\left(k\right)$
and covariance matrix $P_{k|k}^{i}\left(k\right)$; see Appendix \ref{sec:AppendixC}
for details. We should note that, for each Bernoulli, means and covariance
matrices for dead hypotheses ($\overline{x}_{k|k}^{i}\left(l\right)$
and $P_{k|k}^{i}\left(l\right)$ for $l<k$) do not need any further
updating, and multiple hypotheses for the current time are blended
into a single Gaussian.

\subsection{Practical considerations\label{subsec:Practical-considerations}}

In this section, we consider the practical aspects to make the TPMB
filters computationally efficient. As time goes on, the lengths of
the trajectories increase, the sizes of the covariance matrices scale
quadratically and the filtering recursion becomes computationally
demanding. A solution to deal with increasingly long trajectories
is to use an $L$-scan implementation \cite{Angel19_f} in which,
in the prediction step, we approximate the covariance matrices of
the PPP components and the Bernoulli components with the block diagonal
form
\begin{align}
P_{k|k} & \approx\mathrm{diag}\left(\tilde{P}_{k|k}^{t^{k}},\tilde{P}_{k|k}^{t^{k}+1},...,\tilde{P}_{k|k}^{k-L},\tilde{P}_{k|k}^{k-L+1:k}\right)\label{eq:L_scan_cov_approx}
\end{align}
where matrix $\tilde{P}_{k|k}^{k-L+1:k}\in\mathbb{R}^{L\cdot n_{x}\times L\cdot n_{x}}$
represents the joint covariance of the $L$ last time instants, and
$\tilde{P}_{k|k}^{k}\in\mathbb{R}^{n_{x}\times n_{x}}$ represents
the covariance matrix of the target state at time $k$. Thus, states
outside the $L$-scan window are considered independent and remain
unchanged with new measurements. 

The update step requires obtaining the weights for all global hypotheses
$a\in\mathcal{A}_{k}$. In practice, many of these weights are close
to zero and can be pruned before evaluating them by using ellipsoidal
gating and solving the corresponding ranked assignment problem. In
the implementations, we choose the global hypotheses with $N_{h}$
highest weights via Murty's algorithm \cite{Murty68}, in combination
with the Hungarian algorithm, as in \cite{Angel18_b}. A different
approach is to directly approximate the marginal association probabilities
$\overline{w}_{k|k}^{i,a^{i}}$, e.g., as in \cite{Williams14}\cite{Horridge06}.

We also discard Bernoulli components whose existence probability $r_{k|k}^{i}$
is lower than a threshold, and it is also possible to recycle them
\cite{Williams12}. When we consider sets of all trajectories, Bernoulli
components that represent hypotheses of trajectories that have been
detected in the past always have existence probability equal to one.
However, the probability that these trajectories are alive, which
is given by $\beta_{k|k}^{i}\left(k\right)$, can be very low, which
implies that the weights (\ref{eq:detected_weight_all_trajectory})
of their detection hypotheses are very low. Therefore, in order to
avoid computing the weights, means and covariances of these hypotheses,
which have negligible weight when $\beta_{k|k}^{i}\left(k\right)$
is low enough, we set $\beta_{k|k}^{i}\left(k\right)=0$ if $\beta_{k|k}^{i}\left(k\right)$
is less than a threshold $\Gamma_{a}$. In other words, if the probability
that a Bernoulli component that was once detected is alive at the
current time step is lower than $\Gamma_{a}$, it is considered dead,
$\beta_{k|k}^{i}\left(k\right)=0$, which implies that it is no longer
updated or predicted, but it is still a component of the multi-Bernoulli
of the posterior (see (\ref{eq:PMB_k_1_k_1})). In addition, in (\ref{eq:intensity_Gaussian_alive}),
we discard PPP intensity components whose weight is less than a threshold
$\Gamma_{p}$.

\subsection{Estimation\label{subsec:Estimation}}

Given the PMB posterior (\ref{eq:PMB_k_1_k_1}) and a threshold $\Gamma_{d}$,
we use the following computationally efficient estimators. For the
set of alive trajectories, the estimated set of trajectories at time
step $k$ is $\left\{ \left(t^{i},\overline{x}_{k|k}^{i}\right):\,r_{k|k}^{i}>\Gamma_{d}\right\} $,
which reports the starting times and means of the Bernoulli components
whose probability of existence is greater than $\Gamma_{d}$. For
the set of all trajectories, the estimated set of trajectories at
time step $k$ is $\left\{ \left(t^{i},\overline{x}_{k|k}^{i}\left(l^{\star}\right)\right):\,r_{k|k}^{i}>\Gamma_{d},l^{\star}=\underset{l}{\arg\max}\,\beta_{k|k}^{i}\left(l\right)\right\} $,
which reports the starting times and the means with most likely duration
of the Bernoulli components whose probability of existence is greater
than $\Gamma_{d}$. Finally, the pseudocodes of the filters are given
in Algorithm \ref{alg:L-scan_TPMB_algorithms}.

\begin{algorithm}
\caption{\label{alg:L-scan_TPMB_algorithms}Gaussian TPMB filters pseudocode}

{\fontsize{9}{9}\selectfont

\begin{algorithmic}     

\State- Set $\lambda_{0|0}\left(\cdot\right)=0$, $n_{0|0}=0$.

\For{$k=1$ to \textit{final time step} }

\State- Prediction: 

\State$\:$$\circ\,$For alive trajectories: use Lemma \ref{lem:Prediction_Gaussian_TPMB_alive}.

\State$\:$$\circ\,$For all trajectories: use Lemma \ref{lem:Prediction_Gaussian_TPMB_all}.

\State$\:$$\circ\,$Apply $L$-scan approximation (\ref{eq:L_scan_cov_approx})
to all covariance matrices. 

\State- Update:

\State$\:$$\circ\,$For alive trajectories: use Lemma \ref{lem:Update_Gaussian_TPMB_alive}.

\State$\:$$\circ\,$For all trajectories: use Lemma \ref{lem:Update_Gaussian_TPMB_all}.

\State- Use Proposition \ref{lem:Update_Gaussian_TPMB_alive} to
obtain a PMB:

\selectlanguage{british}%
\For{$i=1$ to $n_{k|k}$} 

\selectlanguage{english}%
\State- Calculate $\overline{w}_{k'|k}^{i,a^{i}}$ using (\ref{eq:weight_simplified_Prop2})
and (\ref{eq:weight_global}), and $r_{k|k}^{i}$ using (\ref{eq:existence_Prop2}).

\State- For alive trajectories: calculate $\overline{x}_{k|k}^{i}$
and $P_{k|k}^{i}$ with (\ref{eq:moment_matching_mean}) and (\ref{eq:moment_matching_cov}). 

\State- For all trajectories: calculate $\beta_{k|k}^{i}\left(l\right)$,
$\overline{x}_{k|k}^{i}\left(k\right)$ and $P_{k|k}^{i}\left(k\right)$
with (\ref{eq:beta_KLD_minimisation}), (\ref{eq:mean_KLD_minimisation_all})
and (\ref{eq:cov_KLD_minimisation_all}), see Sec. \ref{subsec:Gaussian-all}.

\EndFor

\State- Estimate the set of trajectories, see Section \ref{subsec:Estimation}. 

\EndFor

\end{algorithmic}

}
\end{algorithm}

\subsection{Discussion\label{subsec:Discussion}}

We proceed to discuss several aspects of the proposed algorithms.
The TPMB filter for alive trajectories has a similar recursion to
the track-oriented PMB filter for sets of targets in \cite{Williams15b},
with the difference that past trajectory states are not integrated
out. For $L=1$, the track-oriented PMB filter and the TPMB filter
perform the same filtering computations, though the TPMB stores the
past means, and possibly the covariances, for each Bernoulli component.
This paper presents the TPMB approximation from direct KLD minimisation
with auxiliary variables. Instead, the derivation in \cite{Williams15b}
uses KLD minimisation on the data association variables, which are
not explicit in the posterior. The TPMB filter for all trajectories
requires the propagation of more variables (\ref{eq:single_trajectory_Gaussian_all}).
A tighter upper bound than (\ref{eq:KLD_bound}) for sets of targets
is studied in \cite{Williams15}.

We have presented the TPMB filters for Poisson birth, as it is generally
more suitable than multi-Bernoulli birth, see Appendix \ref{sec:AppendixD}.
Nevertheless, the above TPMB filter derivation is also valid for multi-Bernoulli
birth. In this case, we just need to set $\lambda_{k'|k}\left(\cdot\right)=0$
and add the Bernoulli components of new born targets in the prediction
step \cite{Angel18_b,Angel19_e}. The resulting algorithm corresponds
to the trajectory multi-Bernoulli (TMB) filter, which can include
target labels to have sets of labelled trajectories, without modifying
the recursion and estimated trajectories \cite{Angel20_c}. We have
presented the Gaussian implementation of the TPMB filter due to its
simplicity and performance. Nevertheless, it is also possible to use
Gaussian mixtures and particles to represent single-trajectory densities. 

Another relevant algorithm is the LMB filter \cite{Reuter14}. As
the $\delta$-GLMB filter, the LMB filter does not work well, unless
practical modifications are used, if there is more than one birth
Bernoulli component with the same mean and covariance \cite[Sec. II.B]{Angel20_b}.
In addition, the LMB update (also the version in \cite{Olofsson17})
makes use of the $\delta$-GLMB update, which requires an exponential
growth in the number of global hypotheses due to the MBM$_{01}$ form
\cite[Sec. IV]{Angel18_b}. The MBM$_{01}$ form is avoided in the
fast LMB in \cite{Kropfreiter20}. The TPMB avoids these drawbacks
by creating Bernoulli components directly from the measurements, performing
the update in PMBM form, without MBM$_{01}$, and estimating trajectories
directly from the posterior. 

\section{Simulations\label{sec:Simulations}}

We analyse the performance of the two TPMB filters\footnote{Matlab code is available at https://github.com/Agarciafernandez.}
in comparison with the trajectory filters: the TPMBM filter \cite{Granstrom18}
, the trajectory global nearest neighbour PMB (TGNPMB) filter, the
TPHD filter and the TCPHD filter \cite{Angel19_f}. The TGNPMB filter
corresponds to the TPMBM filter but only propagating the global hypothesis
with highest weight, as the global nearest neighbor approach \cite{Blackman_book99}.
The TPMB filters have been implemented with the following parameters:
maximum number of hypotheses $N_{h}=200$, threshold for pruning the
PPP weights $\Gamma_{p}=10^{-5}$, threshold for pruning Bernoulli
components $\Gamma_{b}=10^{-5}$ and $L=5$, and $\Gamma_{d}=0.5$
. For the set of all trajectories, we use the same parameters as before
and also parameter $\Gamma_{a}=10^{-4}$. We have also tested the
TMB filter variant in Section \ref{subsec:Discussion} \cite{Angel20_c}.
The TPMBM filters have been implemented with the same parameters and
also with a threshold $10^{-4}$ for pruning global hypothesis weights
and Estimator 1 in \cite{Angel18_b} with threshold 0.4. The TPHD
and TCPHD filters use a pruning threshold $10^{-8}$, absorption threshold
$4$, and limit the number of PHD components to 30 \cite{Angel19_f}. 

The previous trajectory filters are structured to perform smoothing
while filtering in the $L$-scan window. For $L=1$, no single-target
smoothing is performed\foreignlanguage{british}{, though they keep
the probabilistic information on trajectory start time, end time and
past means}. We have also considered three MTT algorithms that do
not exploit trajectory smoothing and estimate the set of trajectories
sequentially by linking the previous estimated trajectories with the
newly estimated targets. The first one is the PMBM filter \cite{Williams15b,Angel18_b}
where the trajectory estimates are formed by linking target state
estimates that originate from the same first detection (same Bernoulli
component).  The PMBM implementation parameters are the same as in
the TPMBM filter. The $\delta$-GLMB filter \cite{Vo17} considers
joint prediction and update using Murty's algorithm, as in \cite{Correa15},
with 1000 global hypotheses and the estimator that first computes
the maximum a posterior of the cardinality \cite{Vo13}. The LMB filter
\cite{Reuter14} also considers 1000 global hypotheses in the update,
maximum number of Gaussians per Bernoulli is 10, the pruning threshold
is $10^{-3}$ for Bernoullis, the pruning threshold is $10^{-5}$
for each Gaussian and the merging threshold for each Gaussian is 4.
The code for $\delta$-GLMB and LMB was obtained from http://ba-tuong.vo-au.com. 

We first consider linear/Gaussian measurements with broad spatial
distribution for new born targets in Section \ref{subsec:Linear_scenario}.
In Section \ref{subsec:Range-bearing-scenario}, we consider range-bearings
measurements with point sources for new born targets.

\subsection{Linear/Gaussian measurements\label{subsec:Linear_scenario}}

We consider a target state $x=\left[p_{x},\dot{p}_{x},p_{y},\dot{p}_{y}\right]^{T}$,
which contains position and velocity in a two-dimensional plane. All
the units in this section are given in the international system. We
use the nearly-constant velocity model with 
\begin{align*}
F=I_{2}\otimes\left(\begin{array}{cc}
1 & \tau\\
0 & 1
\end{array}\right),\quad Q=qI_{2}\otimes\left(\begin{array}{cc}
\tau^{3}/3 & \tau^{2}/2\\
\tau^{2}/2 & \tau
\end{array}\right)
\end{align*}
where $\tau=1$ and $q=0.01$. We also consider $p_{S}=0.99$. The
sensor measures target positions with parameters 
\begin{align*}
H=\left(\begin{array}{cccc}
1 & 0 & 0 & 0\\
0 & 0 & 1 & 0
\end{array}\right),\quad R=\sigma^{2}I_{2},
\end{align*}
where $\sigma^{2}=1$, and $p_{D}=0.9$. The clutter intensity is
$\lambda^{C}\left(z\right)=\overline{\lambda}^{C}u_{A}\left(z\right)$
where $u_{A}\left(z\right)$ is a uniform density in $A=\left[0,300\right]\times\left[0,300\right]$
and $\overline{\lambda}^{C}=10$. The birth intensity is Gaussian
with $\overline{x}_{k}^{b,1}=\left[100,0,100,0\right]^{T}$ and $P_{k}^{b,1}=\mathrm{diag}\left(\left[150^{2},1,150^{2},1\right]\right)$,
with weight $w_{1}^{b,1}=3$ and $w_{k}^{b,1}=0.005$ for $k>1$.

The $\delta$-GLMB filter requires a (labelled) multi-Bernoulli birth
model, not Poisson. For $k>1$, we use one Bernoulli component with
probability of existence 0.005, mean $\overline{x}_{k}^{b,1}$ and
covariance $P_{k}^{b,1}$. For $k=1$, the expected number of targets
is 3 according to the Poisson process, so we consider five Bernoulli
components with the same probability of existence, 0.005, and spatial
distributions such that the support of the multi-Bernoulli birth covers
the expected target number. In this scenario, setting the probability
of existence of the Bernoullis at $k=1$ to 3/5, which makes the multi-Bernoulli
and Poisson process have the same PHD, decreases performance.

This scenario is challenging for the $\delta$-GLMB/LMB filters due
to the fact that there are several potential IID new born targets
with large spatial uncertainty, and the resulting high number of global
hypotheses involved in the first update step, see App. \ref{sec:AppendixD}.
These filters must prune potentially relevant information to be able
to run them in a reasonable time, which implies a loss in performance.
In contrast, at the first time step, the TPMBM and PMBM updates only
require one global hypothesis, which contains full information, is
already in PMB form and is very fast to compute. Due to the Gaussian
implementation with moment matching, the TMB filter estimates trajectories
with coalescence from the beginning, so it is not considered further.

\begin{figure}
\begin{centering}
\includegraphics[scale=0.32]{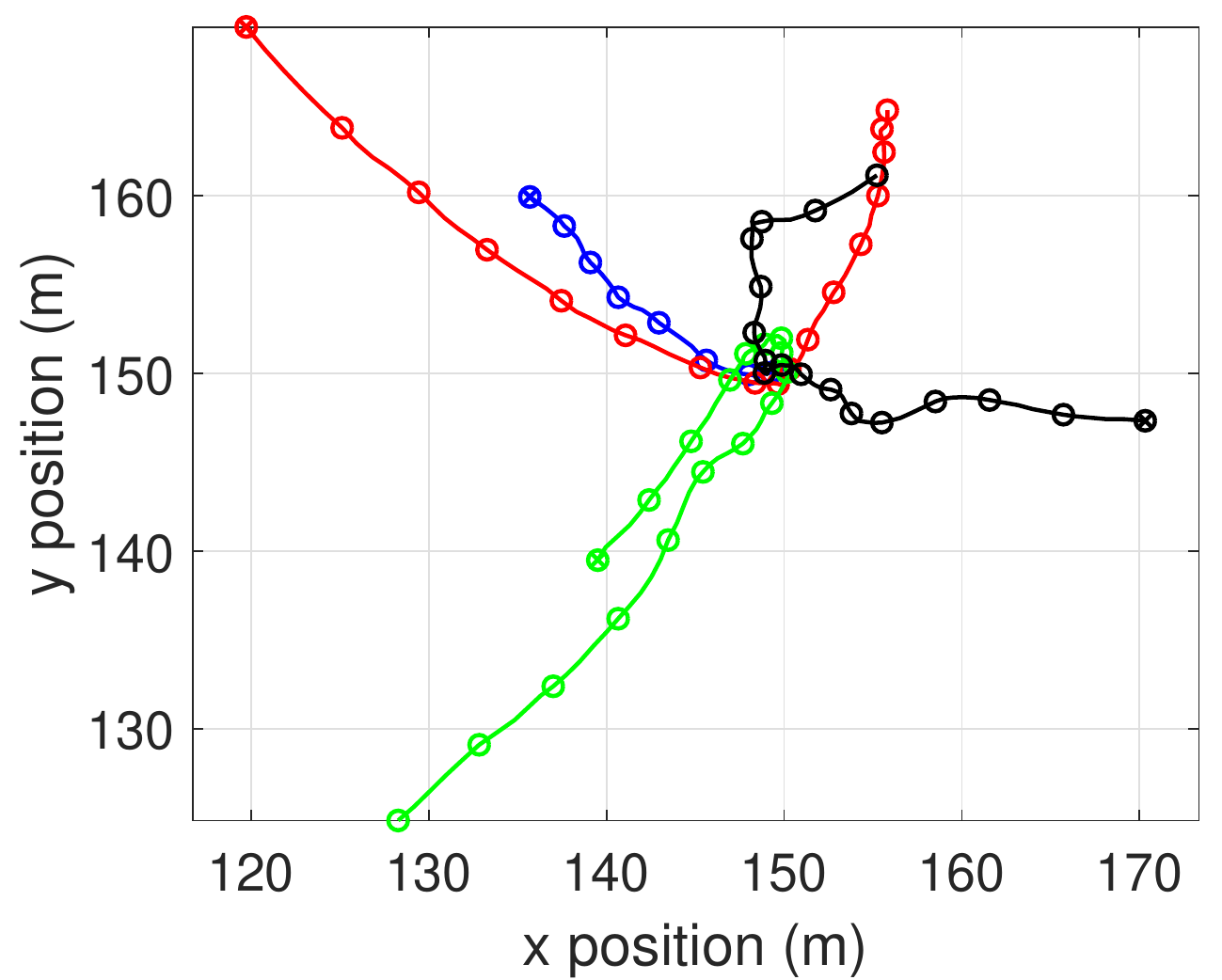}\includegraphics[scale=0.32]{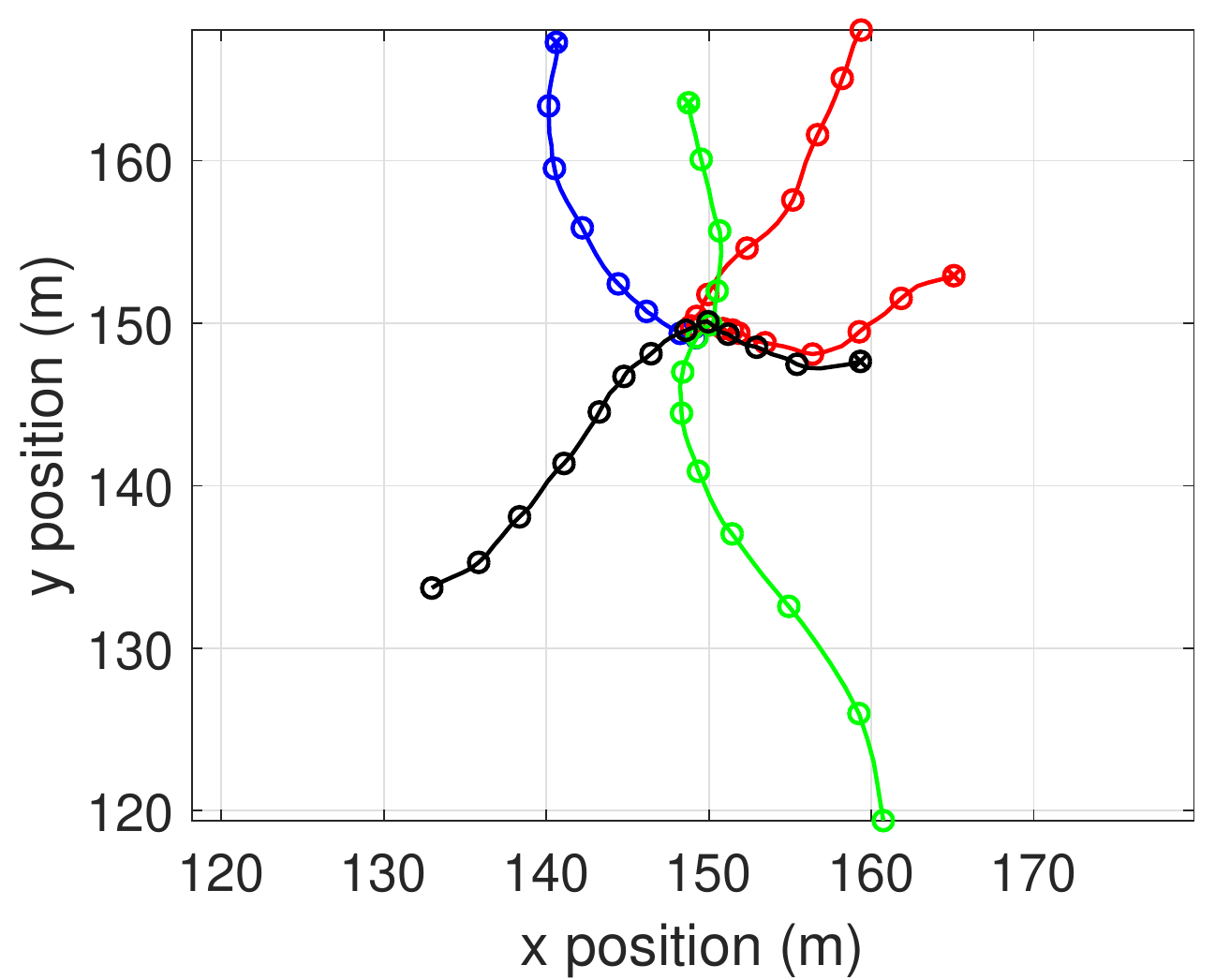}
\par\end{centering}
\caption{\label{fig:Scenario}True trajectories in Scenario 1 (left) \cite{Angel18_b}
and Scenario 2 (right). In Scenario 1, all targets are born at time
step 1. In Scenario 2, the blue and the red targets are born at time
step 1, whereas the green and the black targets are born at time step
21. The only target that dies during the simulations is the blue target,
which dies at time step 40, when all targets are in close proximity.
Targets positions every 10 time steps are marked with a circle, and
their initial positions with a filled circle.}
\end{figure}

We consider the ground truth trajectories with $N_{s}=81$ time steps
in Figure \ref{fig:Scenario}(left). We assess filter performance
using Monte Carlo simulation with $N_{mc}=100$ runs. At each time
step $k$, we measure the error between the true set $\mathbf{X}_{k}$
of trajectories and its estimate $\mathbf{\hat{X}}_{k}$, which differ
depending on the problem formulation (see Section \ref{sec:Problem-formulation}).
The error is determined by the linear programming metric $d\left(\cdot,\cdot\right)$
for sets of trajectories in \cite{Angel20_d} with parameters $p=2$,
$c=10$ and $\gamma=1$. In our results, we only use the position
elements to compute $d\left(\cdot,\cdot\right)$ and normalise the
error by the considered time window such that the squared error at
time $k$ becomes $d^{2}\left(\mathbf{X}_{k},\mathbf{\hat{X}}_{k}\right)/k$.
The root mean square (RMS) error at a given time step is
\begin{align}
d\left(k\right) & =\sqrt{\frac{1}{N_{mc}k}\sum_{i=1}^{N_{mc}}d^{2}\left(\mathbf{X}_{k},\mathbf{\hat{X}}_{k}^{i}\right)},\label{eq:error_time_k}
\end{align}
where $\mathbf{\hat{X}}_{k}^{i}$ is the estimate of the set of trajectories
at time $k$ in the $i$th Monte Carlo run. 

\begin{figure}
\begin{centering}
\includegraphics[scale=0.6]{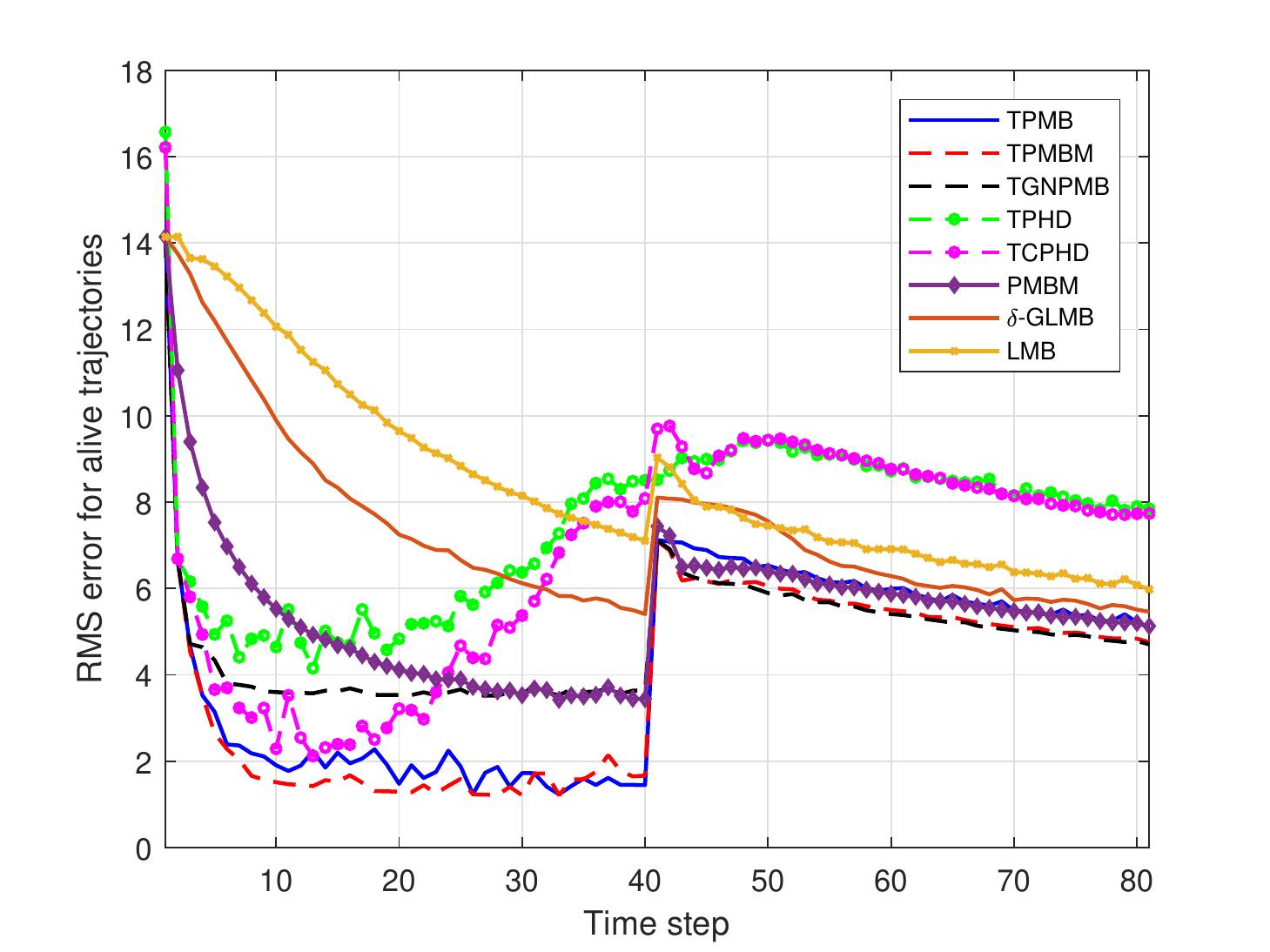}
\par\end{centering}
\caption{\label{fig:Trajectory-metric-error-alive}Trajectory metric error
against time for the alive trajectories. Error increases at time step
40, when a target dies. On the whole, the TPMBM filter is the best
performing filter followed by the TPMB filter. }
\end{figure}

We first proceed to analyse the error in estimating the set of alive
trajectories. The RMS trajectory error against time is shown in Figure
\ref{fig:Trajectory-metric-error-alive}. For all filters, estimation
error increases after time step 40, in which a target dies, and the
rest of the targets are in close proximity. The TPMBM filter is the
most accurate filter to estimate the alive trajectories. This is to
be expected, as without approximations, the TPMBM filtering recursion
provides the true posterior over the set of trajectories. The second
best performing filter in general is the proposed TPMB filter. Its
performance is slightly worse than TGNPMB after time step 40 but it
is considerably better before time step 40. TPHD and TCPHD filters
perform considerably worse, as they are less accurate approximations.
The LMB filter performs worse than $\delta$-GLMB. $\delta$-GLMB
is less accurate than PMBM, which is outperformed by TPMBM and TPMB. 

\begin{figure}
\begin{centering}
\includegraphics[scale=0.3]{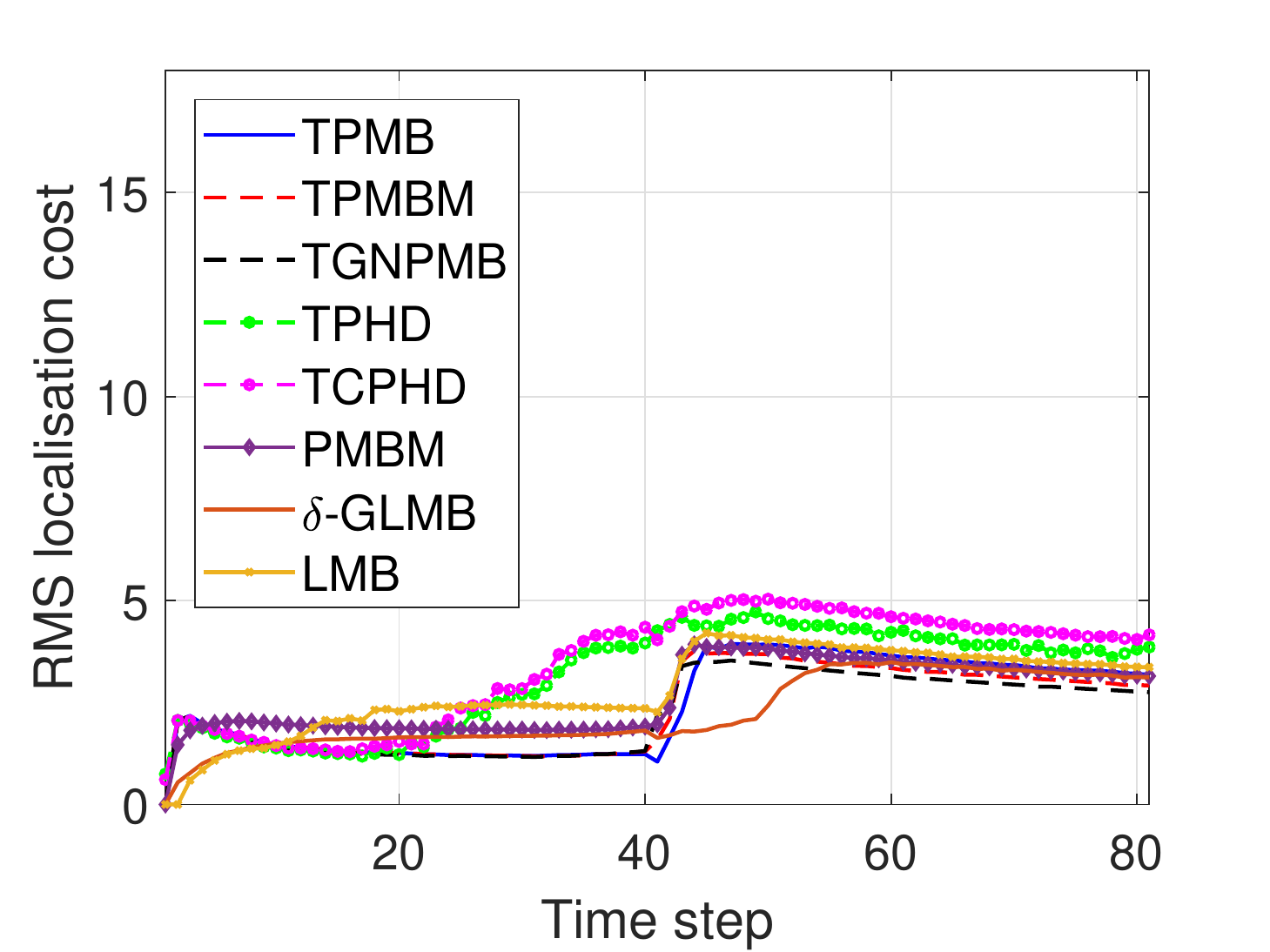}\includegraphics[scale=0.3]{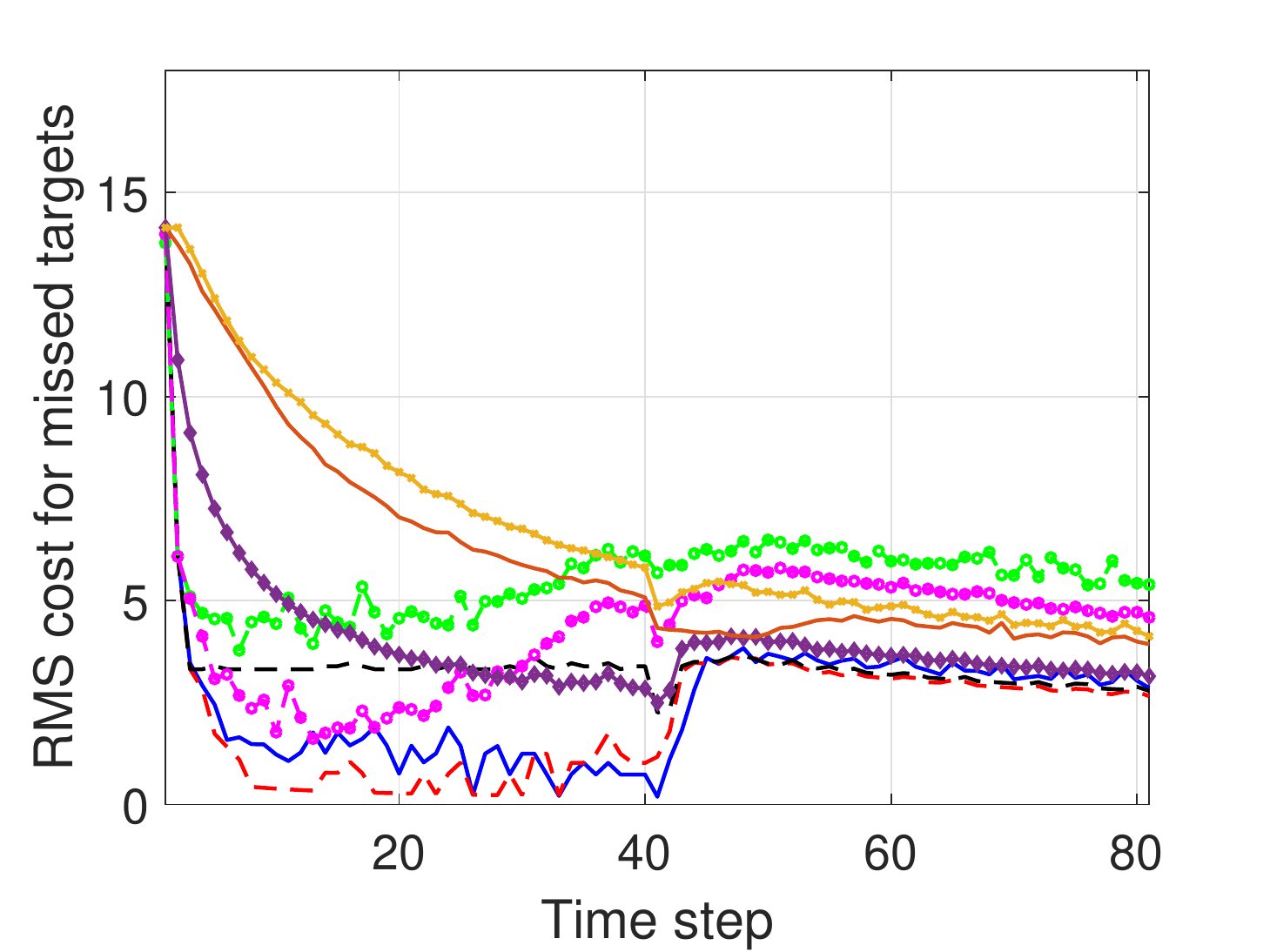}
\par\end{centering}
\begin{centering}
\includegraphics[scale=0.3]{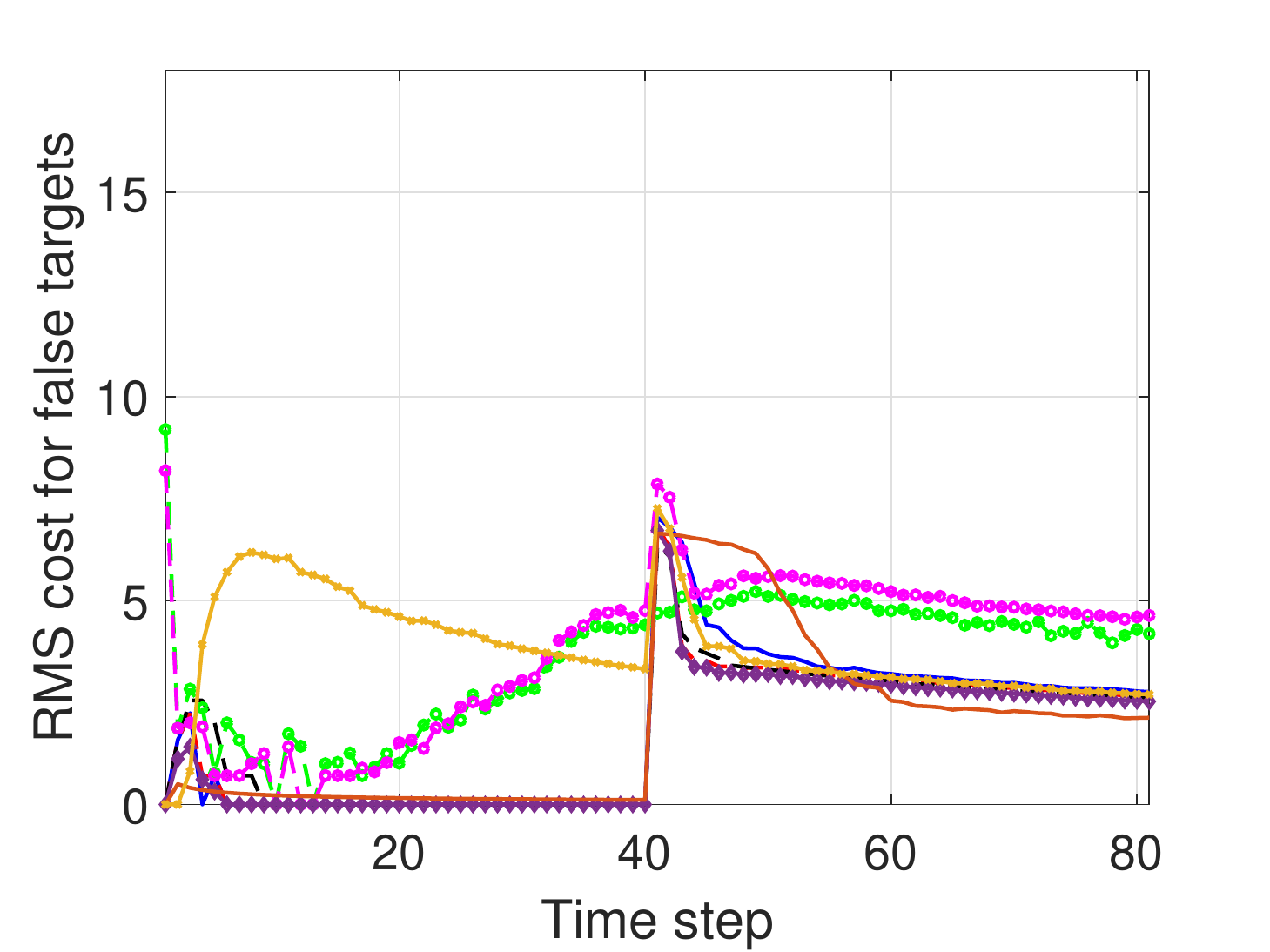}\includegraphics[scale=0.3]{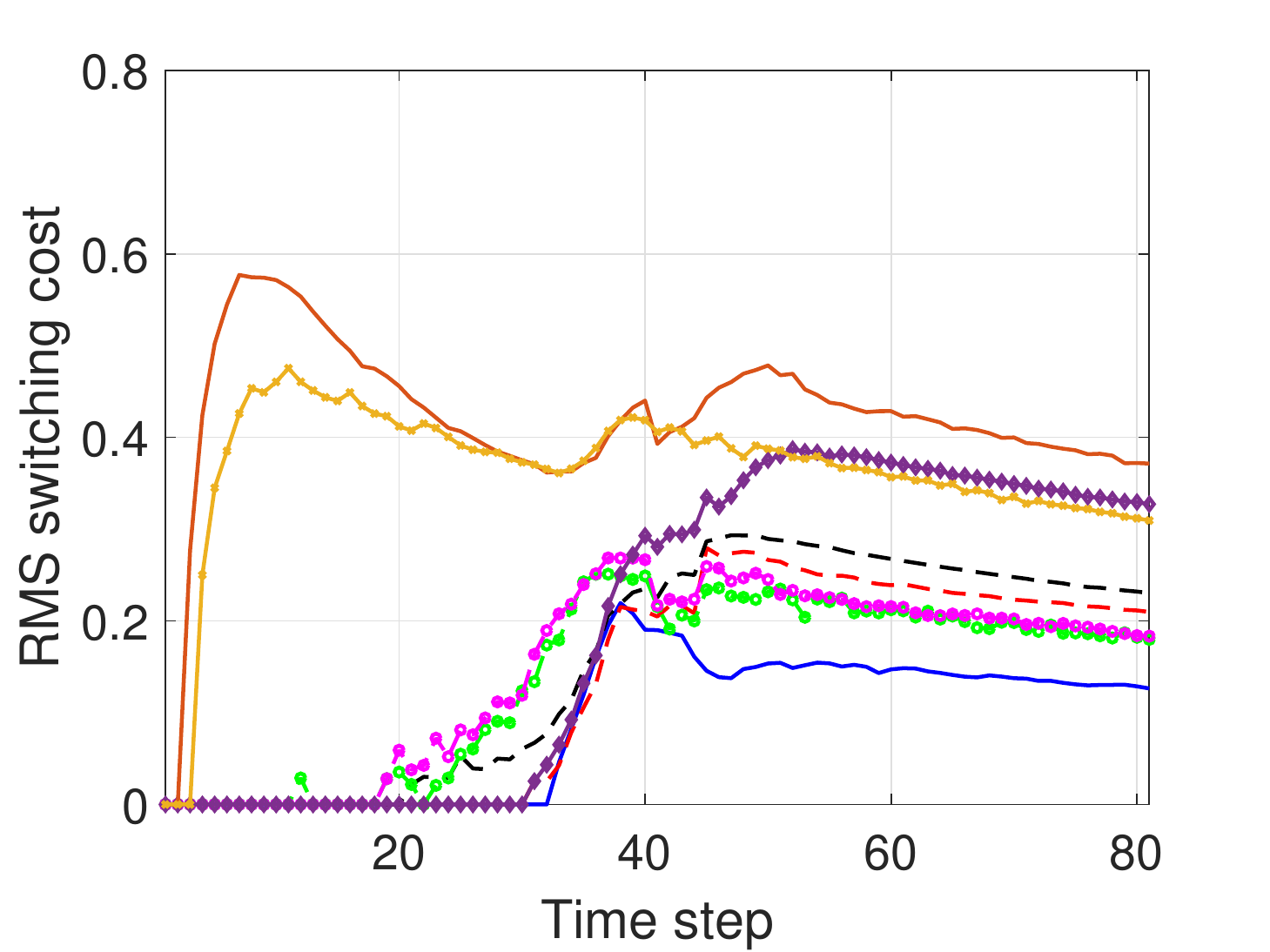}
\par\end{centering}
\caption{\label{fig:Decomposition-error_alive}Decomposition of the trajectory
metric error against time for the alive trajectories.}
\end{figure}

The squared trajectory metric $d^{2}\left(\cdot,\cdot\right)$ can
be decomposed into the square costs for missed targets, false targets,
localisation error of properly detected targets, and track switches
\cite{Angel20_d}. The resulting RMS errors for the decomposed costs
are shown in Figure \ref{fig:Decomposition-error_alive}. Before time
step 40, all filters except LMB, TPHD and TCPHD have a negligible
cost for false targets. Also, $\delta$-GLMB and PMBM show a higher
error for missed targets, followed by TPHD, TCPHD and TGNPMBM. After
time step 40, these filters increase their error mainly due to false
and missed target errors, created by the disappearance of one trajectory.
Track switching cost are small and quite similar for all filters based
on sets of trajectories. $\delta$-GLMB and LMB are the only filters
with track switches before targets get in close proximity, due to
the IID birth, and provide the highest switching costs. PMBM filter
has the third highest switching costs after time step 40. The errors
obtained by the square sum of the generalised optimal sub-pattern
assignment (GOSPA) metric ($\alpha=2$) \cite{Rahmathullah17} at
each time step instead of $d^{2}\left(\cdot,\cdot\right)$ in (\ref{eq:error_time_k})
are quite similar to the trajectory metric errors, as, due to the
choice of $\gamma$, the switching costs are small. 

The average execution times in seconds of a single run (81 time steps)
of our Matlab implementations with a 3.5 GHz Intel Xeon E5 processor
are: 1.2 (TPMB), 7.0 (TPMBM), 0.7 (TGNPMB), 1.1 (TPHD), 1.1 (TCPHD),
5.8 (PMBM), 13.1 ($\delta$-GLMB) and 10.5 (LMB).  The fastest algorithm
is TGNPMB, though its performance is worse than TPMB. The TPMBM is
slower than the other trajectory filters and PMBM but TPMBM is also
the one with highest performance, as expected. There is a trade-off
between computational complexity and accuracy in the selection of
TPMBM, TPMB and TGNPMB. Trajectory filters are faster than $\delta$-GLMB/LMB
even though the update past trajectory information in the $L$-scan
window due to the considerably lower number of required global hypotheses
to keep relevant information, see App. \ref{sec:AppendixD}.

We proceed to analyse the performance of the filters for different
values of $L$ and different $p_{D}$ and $\overline{\lambda}^{C}$.
In Table \ref{tab:Trajectory-metric-errors-alive}, we show the resulting
RMS error considering all time steps
\begin{align}
d_{T}= & \sqrt{\frac{1}{N_{s}}\sum_{k=1}^{N_{s}}d^{2}\left(k\right)}.\label{eq:RMS_error_all_time}
\end{align}
Increasing $L$ for the trajectory filters lowers the error, mainly
due to improved localisation of past states. In general, the best
performing filter is TPMBM, followed by TPMB. \foreignlanguage{british}{The
TPMB approximation is accurate for the considered probabilities of
detection and clutter intensity.} As expected, if the clutter intensity
increases, performance decreases for all filters. The higher the probability
of detection, performance increases. 

\begin{table*}
\caption{\label{tab:Trajectory-metric-errors-alive}Trajectory metric errors
(alive trajectories)}

\centering{}%
\begin{tabular}{c|>{\centering}p{0.4cm}>{\centering}p{0.4cm}>{\centering}p{0.4cm}|>{\centering}p{0.4cm}>{\centering}p{0.4cm}>{\centering}p{0.4cm}|>{\centering}p{0.4cm}>{\centering}p{0.4cm}>{\centering}p{0.4cm}|>{\centering}p{0.4cm}>{\centering}p{0.4cm}>{\centering}p{0.4cm}|>{\centering}p{0.4cm}>{\centering}p{0.4cm}>{\centering}p{0.4cm}||>{\centering}p{0.7cm}|>{\centering}p{0.7cm}|>{\centering}p{0.7cm}}
\hline 
 &
\multicolumn{3}{c|}{TPMB} &
\multicolumn{3}{c|}{TPMBM} &
\multicolumn{3}{c|}{TGNPMB} &
\multicolumn{3}{c|}{TPHD} &
\multicolumn{3}{c||}{TCPHD} &
PMBM  &
GLMB &
LMB\tabularnewline
\hline 
$L$ &
1 &
5 &
10 &
1 &
5 &
10 &
1 &
5 &
10 &
1 &
5 &
10 &
1 &
5 &
10 &
- &
- &
-\tabularnewline
\hline 
No change &
5.18 &
4.84 &
4.82 &
4.87 &
4.50 &
4.49 &
5.30 &
4.98 &
4.97 &
7.87 &
7.69 &
7.69 &
7.50 &
7.30 &
7.30 &
5.73 &
7.62 &
8.64\tabularnewline
$p_{D}=0.99$  &
4.91 &
4.73 &
4.73 &
4.55 &
4.36 &
4.35 &
4.66 &
4.47 &
4.46 &
7.17 &
7.06 &
7.06 &
7.07 &
6.96 &
6.96 &
5.46 &
7.38 &
7.48\tabularnewline
$p_{D}=0.80$  &
5.52 &
5.02 &
5.00 &
5.38 &
4.89 &
4.88 &
5.94 &
5.54 &
5.52 &
8.74 &
8.52 &
8.52 &
8.04 &
7.80 &
7.79 &
6.27 &
7.52 &
9.48\tabularnewline
$p_{D}=0.70$  &
5.92 &
5.21 &
5.18 &
5.91 &
5.24 &
5.22 &
7.02 &
6.55 &
6.53 &
9.12 &
8.88 &
8.87 &
8.47 &
8.20 &
8.20 &
6.76 &
8.43 &
10.38\tabularnewline
$\overline{\lambda}^{C}=20$ &
5.27 &
4.92 &
4.91 &
4.98 &
4.63 &
4.62 &
6.39 &
6.18 &
6.17 &
8.04 &
7.87 &
7.86 &
7.55 &
7.36 &
7.36 &
5.90 &
8.19 &
9.12\tabularnewline
$\overline{\lambda}^{C}=30$ &
5.27 &
4.96 &
4.94 &
4.95 &
4.62 &
4.61 &
7.50 &
7.39 &
7.39 &
8.11 &
7.95 &
7.95 &
7.62 &
7.44 &
7.44 &
5.96 &
8.51 &
9.44\tabularnewline
$\overline{\lambda}^{C}=40$ &
5.32 &
5.01 &
4.99 &
5.05 &
4.73 &
4.72 &
8.14 &
8.04 &
8.04 &
8.16 &
8.01 &
8.00 &
7.73 &
7.56 &
7.56 &
6.12 &
8.86 &
9.79\tabularnewline
\hline 
\end{tabular}
\end{table*}

Finally, we consider the estimation of the set of all trajectories.
TPHD and TCPHD filters are not included as they are not suitable for
this problem \cite{Angel19_f}. The RMS errors (\ref{eq:RMS_error_all_time})
are shown in Table \ref{tab:Trajectory-metric-errors-all}. As before,
error decreases by increasing $L$, and TPMBM is generally the best
algorithm followed by TPMB. PMBM performs worse than these filters,
but better than $\delta$-GLMB and LMB.

The average execution times in seconds of a single run (81 time steps)
for $p_{D}=0.9$, $\overline{\lambda}^{C}=10$, $L=5$, and the estimation
of all trajectories are: 1.5 (TPMB), 7.9 (TPMBM), 0.9 (TGNPMB), 5.8
(PMBM), 13.1 ($\delta$-GLMB) and 10.5 (LMB). Compared to tracking
alive trajectories, there is an increase in the execution time in
the trajectory filters. The sequential track estimators, PMBM, $\delta$-GLMB
and LMB have the same computational burden to solve both problems,
as they only differ in the estimated set of trajectories.

\begin{table*}
\caption{\label{tab:Trajectory-metric-errors-all}Trajectory metric errors
(all trajectories)}

\centering{}%
\begin{tabular}{c|ccc|ccc|ccc||c|c|c}
\hline 
 &
\multicolumn{3}{c|}{TPMB} &
\multicolumn{3}{c|}{TPMBM} &
\multicolumn{3}{c||}{TGNPMB} &
PMBM &
GLMB &
LMB\tabularnewline
\hline 
$L$ &
1 &
5 &
10 &
1 &
5 &
10 &
1 &
5 &
10 &
- &
- &
-\tabularnewline
\hline 
No change &
3.11 &
2.37 &
2.34 &
3.09 &
2.36 &
2.34 &
4.31 &
3.88 &
3.86 &
4.58 &
7.05 &
8.79\tabularnewline
$p_{D}=0.99$  &
2.52 &
2.07 &
2.04 &
2.50 &
2.06 &
2.04 &
3.11 &
2.78 &
2.77 &
4.17 &
6.68 &
6.74\tabularnewline
$p_{D}=0.80$  &
3.71 &
2.77 &
2.73 &
3.66 &
2.75 &
2.72 &
5.22 &
4.70 &
4.68 &
5.15 &
7.52 &
9.63\tabularnewline
$p_{D}=0.70$  &
4.43 &
3.26 &
3.20 &
4.41 &
3.29 &
3.25 &
6.56 &
6.00 &
5.98 &
5.70 &
8.02 &
10.84\tabularnewline
$\overline{\lambda}^{C}=20$ &
3.23 &
2.52 &
2.49 &
3.18 &
2.49 &
2.47 &
6.22 &
5.99 &
5.99 &
4.81 &
7.75 &
9.07\tabularnewline
$\overline{\lambda}^{C}=30$ &
3.29 &
2.64 &
2.61 &
3.24 &
2.63 &
2.60 &
7.63 &
7.53 &
7.52 &
5.02 &
8.20 &
9.52\tabularnewline
$\overline{\lambda}^{C}=40$ &
3.31 &
2.69 &
2.66 &
3.29 &
2.70 &
2.68 &
8.39 &
8.30 &
8.30 &
5.17 &
8.69 &
9.97\tabularnewline
\hline 
\end{tabular}
\end{table*}

\subsection{Range-bearings measurements\label{subsec:Range-bearing-scenario}}

This section analyses a scenario with range-bearings measurements
\cite{Ristic12}, and the same dynamic model as in Section \ref{subsec:Linear_scenario}.
We have $l\left(z|x\right)=\mathcal{N}\left(z;h\left(x\right),R\right)$
with
\begin{align*}
h\left(x\right) & =\left[\sqrt{\left(p_{x}-s_{x}\right)^{2}+\left(p_{y}-s_{y}\right)^{2}},\arctan\left(\frac{p_{y}-s_{y}}{p_{x}-s_{x}}\right)\right]^{T}
\end{align*}
where $\left[s_{x},s_{y}\right]=\left[100,100\right]$ is the sensor
location, and $R=\text{diag}([1,\left(2\pi/180\right)^{2}])$.

We consider multi-Bernoulli birth with four Bernoullis with Gaussian
densities located at point sources \cite{Angel19_e}. The probabilities
of existence are 0.01, the covariance matrices are $\text{diag}([9,1,9,1])$,
and the means are located at $[140,0,170,0]^{T}$, $[165,0,155,0]^{T}$,
$[150,0,160,0]^{T}$ and $[160,0,150,0]^{T}$, respectively. The
filters with Poisson RFS birth model have an intensity that matches
the PHD, which is a Gaussian mixture. We consider $p_{D}=0.9$ and
$\lambda^{C}\left(z\right)=\overline{\lambda}^{C}u_{A}\left(z\right)$
where $A=\left[10,200\right]\times\left[0,\pi/2\right]$ and $\overline{\lambda}^{C}=10$. 

We have implemented the filtering recursions using an extended Kalman
filter (EKF) \cite{Sarkka_book13}. The EKF linearises $h\left(\cdot\right)$
at the current mean of each trajectory density using a first-order
Taylor series. Then, the multi-target filtering recursions proceed
as in the affine measurement case, which is a direct extension of
the linear case.

\begin{figure}
\begin{centering}
\includegraphics[scale=0.6]{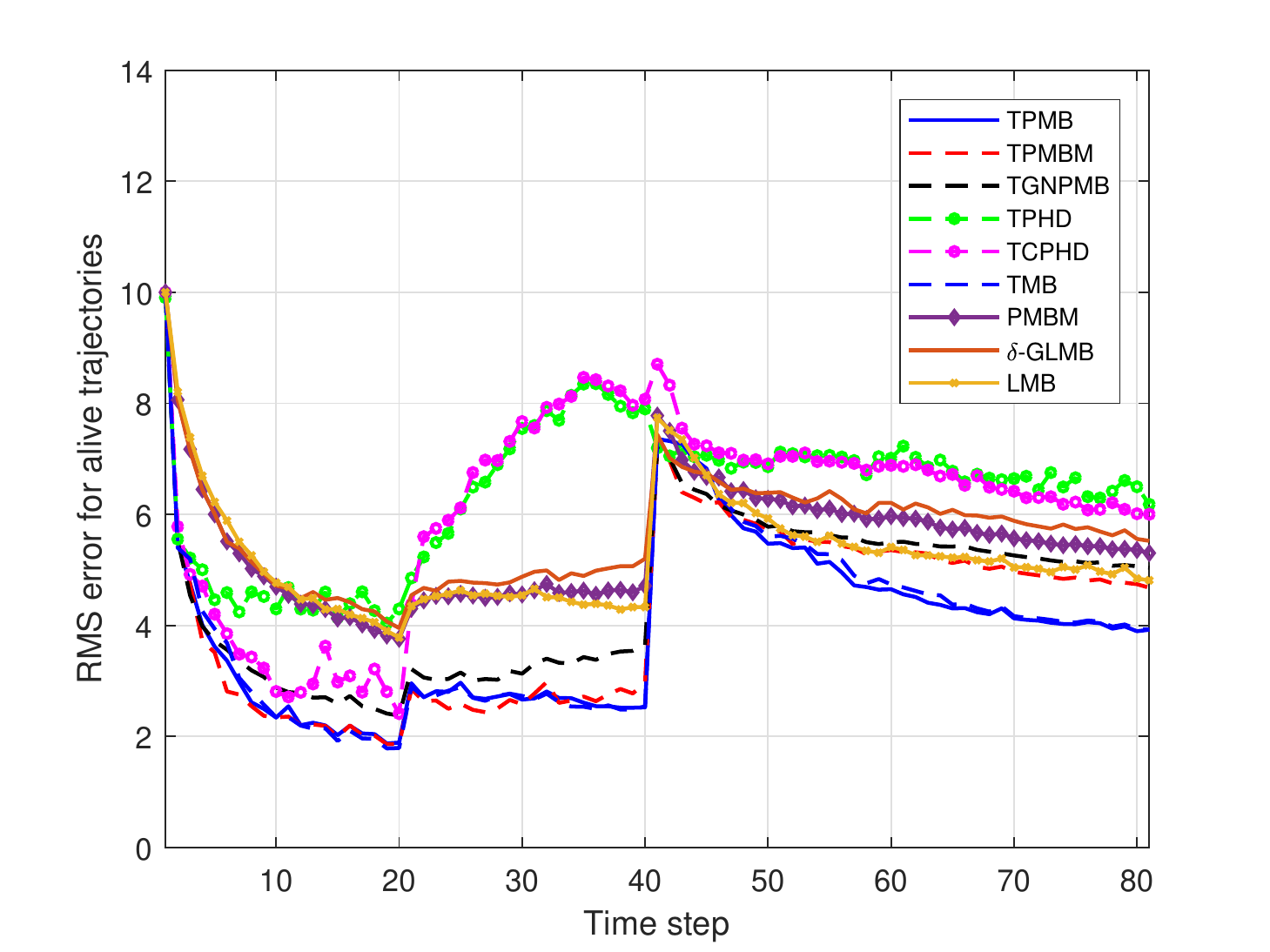}
\par\end{centering}
\caption{\label{fig:Trajectory-metric-error_range_bearings}Trajectory metric
error against time for the alive trajectories in the range-bearings
scenario. The TPMB and TMB filters are the best performing filters,
with TPMB being faster than TMB.}

\end{figure}

The ground truth set of trajectories, with $81$ time steps, is shown
in Figure \ref{fig:Scenario} (right). The trajectory filters are
implemented with $L=5$. The RMS trajectory error for alive trajectories
is shown in Figure \ref{fig:Trajectory-metric-error_range_bearings}.
In this case, the TPMB and TMB filters are the best performing filters
followed by the TPMBM filter and TGNPMB. Sequential track estimators,
PMBM and $\delta$-GLMB perform quite similarly, and LMB works slightly
better than these for the considered parameters. TPHD and TCPHD have
lower performance. 

The average execution times in seconds (81 time steps) are: 1.8 (TPMB),
5.9 (TPMBM), 0.6 (TGNPMB), 0.9 (TPHD), 0.9 (TCPHD), 5.5 (TMB), 4.9
(PMBM), 14.8 ($\delta$-GLMB) and 30.1 (LMB).  The filters with Poisson
birth are faster than with multi-Bernoulli birth. The joint prediction
and update $\delta$-GLMB is faster than LMB \cite{Reuter14} in this
scenario. 

\section{Conclusions\label{sec:Conclusions}}

We have proposed two Poisson multi-Bernoulli filters for sets of trajectories
to perform multiple target tracking. One TPMB filter contains information
on the set of alive trajectories and the other on the set of all trajectories,
which include alive and dead trajectories.  We have also proposed
Gaussian implementations of the filters. 

The resulting filters offer a trade-off between computational complexity
and accuracy. They are faster than TPMBM filters but typically have
worse performance. The TPMB filters are considerably more accurate
than TPHD and TCPHD filters, but with higher computational complexity.
Computational and performance benefits with respect to $\delta$-GLMB
and LMB filters are shown in two simulated scenarios. 

\bibliographystyle{IEEEtran}
\bibliography{9C__Trabajo_laptop_Mis_articulos_Finished_Trajectory_PMB_filter_Accepted_Referencias}

\cleardoublepage{}

{\LARGE{}Trajectory Poisson multi-Bernoulli filters: Supplementary
material}{\LARGE\par}

\appendices{}

\section{\label{sec:AppendixA}}

In this appendix, we prove that if we integrate the auxiliary variables
in $\widetilde{f}_{k'|k}\left(\cdot\right)$, see (\ref{eq:PMBM_aux_var2}),
we recover $f_{k'|k}\left(\cdot\right)$. That is, we prove that
\begin{align}
 & \sum_{u_{1:n}\in\mathbb{\mathbb{U}}_{k}^{n}}\widetilde{f}_{k'|k}\left(\left\{ \left(u_{1},X_{1}\right),...,\left(u_{n},X_{n}\right)\right\} \right)\nonumber \\
 & \quad=f_{k'|k}\left(\left\{ X_{1},...,X_{n}\right\} \right).\label{eq:integrate_out_auxiliary}
\end{align}
We first obtain two preliminary results with only a Poisson component
and a multi-Bernoulli mixture. Then, we proceed to prove the PMBM
case.

\subsection{PPP}

Integrating out the auxiliary variables in $\widetilde{f}_{k'|k}^{p}\left(\cdot\right)$,
we obtain
\begin{align}
 & \sum_{u_{1:n}\in\mathbb{\mathbb{U}}_{k'|k}^{n}}\widetilde{f}_{k'|k}^{p}\left(\left\{ \left(u_{1},X_{1}\right)...,\left(u_{n},X_{n}\right)\right\} \right)\nonumber \\
 & =e^{-\int\lambda_{k'|k}\left(X\right)dX}\left[\prod_{i=1}^{n}\lambda_{k'|k}\left(X_{i}\right)\right]\left[\sum_{u_{1:n}\in\mathbb{\mathbb{U}}_{k'|k}^{n}}\prod_{i=1}^{n}\delta_{0}\left[u_{i}\right]\right]\nonumber \\
 & =f_{k'|k}^{p}\left(\left\{ X_{1},...,X_{n}\right\} \right).\label{eq:integrate_out_PPP}
\end{align}

\subsection{Multi-Bernoulli mixture}

We first note that if $\lambda_{k'|k}\left(\cdot\right)=0$, then,
the PMBM (\ref{eq:TPMBM_original}) is an MBM. Using \cite[Eq. (4.127)]{Mahler_book14},
for existence probabilities smaller than one, the MBM with auxiliary
variables can be written as
\begin{align}
 & \widetilde{f}_{k'|k}^{mbm}\left(\left\{ \left(u_{1},X_{1}\right)...,\left(u_{n},X_{n}\right)\right\} \right)\nonumber \\
 & =\sum_{a\in\mathcal{A}_{k'|k}}w_{k'|k}^{a}\left[\prod_{i=1}^{n_{k'|k}}\left(1-r_{k'|k}^{i,a^{i}}\right)\right]\nonumber \\
 & \sum_{1\leq i_{1}\neq...,\neq i_{n}\leq n_{k'|k}}\prod_{p=1}^{n}\frac{r_{k'|k}^{i_{p},a_{p}^{i}}}{1-r_{k'|k}^{i_{p},a_{p}^{i}}}p_{k'|k}^{i_{p},a^{i_{p}}}\left(X_{p}\right)\delta_{i_{p}}\left[u_{p}\right]\label{eq:MBM_aux_append}
\end{align}
If we integrate out the auxiliary variables in (\ref{eq:MBM_aux_append}),
we obtain the MBM without auxiliary variables

\begin{align}
 & \sum_{u_{1:n}\in\mathbb{\mathbb{U}}_{k'|k}^{n}}\widetilde{f}_{k'|k}^{mbm}\left(\left\{ \left(u_{1},X_{1}\right)...,\left(u_{n},X_{n}\right)\right\} \right)\nonumber \\
 & =\sum_{a\in\mathcal{A}_{k'|k}}w_{k'|k}^{a}\left[\prod_{i=1}^{n_{k'|k}}\left(1-r_{k'|k}^{i,a^{i}}\right)\right]\sum_{1\leq i_{1}\neq...,\neq i_{n}\leq n_{k'|k}}\nonumber \\
 & \,\left[\prod_{p=1}^{n}\frac{r_{k'|k}^{i_{p},a_{p}^{i}}}{1-r_{k'|k}^{i_{p},a_{p}^{i}}}p_{k'|k}^{i_{p},a^{i_{p}}}\left(X_{p}\right)\right]\left[\sum_{u_{1:n}\in\mathbb{\mathbb{U}}_{k'|k}^{n}}\prod_{p=1}^{n}\delta_{i_{p}}\left[u_{p}\right]\right]\\
 & =f_{k'|k}^{mbm}\left(\left\{ X_{1}...,X_{n}\right\} \right).
\end{align}

If some existence probabilities are equal to one, the derivation is
analogous, but removing the corresponding $\left(1-r_{k'|k}^{i,a^{i}}\right)$
in the numerator and denominator in (\ref{eq:MBM_aux_append}). 

\subsection{PMBM}

We can write the PMBM (\ref{eq:TPMBM_original}) as
\begin{align}
f_{k'|k}\left(\left\{ X_{1},...,X_{n}\right\} \right) & =\sum_{\mathbf{Y}\subseteq\left\{ X_{1},...,X_{n}\right\} }f_{k'|k}^{p}\left(\mathbf{Y}\right)f_{k'|k}^{mbm}\left(\mathbf{X}_{k'}\setminus\mathbf{Y}\right)\\
 & =\sum_{j=0}^{n}\sum_{\sigma\in\Gamma_{n,j}}f_{k'|k}^{p}\left(\left\{ X_{\sigma_{1}},...,X_{\sigma_{j}}\right\} \right)\nonumber \\
 & \quad\times f_{k'|k}^{mbm}\left(\left\{ X_{\sigma_{j+1}},...,X_{\sigma_{n}}\right\} \right)
\end{align}
 where $\Gamma_{n,j}$ is the set that contains all possible sets
$\sigma=\left\{ \sigma_{1},...,\sigma_{j}\right\} $ of $j$ elements
from $\left\{ 1,...,n\right\} $ and $\left\{ \sigma_{j+1},...,\sigma_{n}\right\} =\left\{ 1,...,n\right\} \setminus\left\{ \sigma_{1},...,\sigma_{j}\right\} $.
The cardinality of set $\Gamma_{n,j}$ is
\begin{equation}
\left|\Gamma_{n,j}\right|=\left(\begin{array}{c}
n\\
j
\end{array}\right).
\end{equation}

Using these formulas on the PMBM density with auxiliary variables,
we have
\begin{align}
 & \widetilde{f}_{k'|k}\left(\left\{ \left(u_{1},X_{1}\right)...,\left(u_{n},X_{n}\right)\right\} \right)\nonumber \\
 & =\sum_{j=0}^{n}\sum_{\sigma\in\Gamma_{n,j}}\widetilde{f}_{k'|k}^{p}\left(\left\{ \left(u_{\sigma_{1}},X_{\sigma_{1}}\right),...,\left(u_{\sigma_{j}},X_{\sigma_{j}}\right)\right\} \right)\nonumber \\
 & \times\widetilde{f}_{k'|k}^{mbm}\left(\left\{ \left(u_{\sigma_{j+1}},X_{\sigma_{j+1}}\right),...,\left(u_{\sigma_{n}},X_{\sigma_{n}}\right)\right\} \right)
\end{align}
Then, integrating out the auxiliary variables
\begin{align}
 & \sum_{u_{1:n}\in\mathbb{\mathbb{U}}_{k'|k}^{n}}\widetilde{f}_{k'|k}\left(\left\{ \left(u_{1},X_{1}\right)...,\left(u_{n},X_{n}\right)\right\} \right)\nonumber \\
 & =\sum_{u_{1:n}\in\mathbb{\mathbb{U}}_{k'|k}^{n}}\sum_{j=0}^{n}\sum_{\sigma\in\Gamma_{n,j}}\widetilde{f}_{k'|k}^{p}\left(\left\{ \left(u_{\sigma_{1}},X_{\sigma_{1}}\right),...,\left(u_{\sigma_{j}},X_{\sigma_{j}}\right)\right\} \right)\nonumber \\
 & \times\widetilde{f}_{k'|k}^{mbm}\left(\left\{ \left(u_{\sigma_{j+1}},X_{\sigma_{j+1}}\right),...,\left(u_{\sigma_{n}},X_{\sigma_{n}}\right)\right\} \right)\\
 & =\sum_{j=0}^{n}\sum_{\sigma\in\Gamma_{n,j}}\left[\sum_{u_{\sigma_{1:j}}\in\mathbb{\mathbb{N}}_{0}^{j}}\widetilde{f}_{k'|k}^{p}\left(\left\{ \left(u_{\sigma_{1}},X_{\sigma_{1}}\right),...,\left(u_{\sigma_{j}},X_{\sigma_{j}}\right)\right\} \right)\right]\nonumber \\
 & \times\left[\sum_{u_{\sigma_{j+1:n}}\in\mathbb{\mathbb{N}}_{0}^{n-j}}\widetilde{f}_{k'|k}^{mbm}\left(\left\{ \left(u_{\sigma_{j+1}},X_{\sigma_{j+1}}\right),...,\left(u_{\sigma_{n}},X_{\sigma_{n}}\right)\right\} \right)\right]
\end{align}
Applying the results in the previous two subsections, we finish the
proof of (\ref{eq:integrate_out_auxiliary}).

\section{\label{sec:AppendixB}}

In Section \ref{subsec:app_KLD-minimisation}, we prove Proposition
\ref{prop:PMB_KLD_minimisation}. We also prove in Section \ref{subsec:Matching-the-PHD}
that the resulting density from the KLD minimisation also matches
the PHD. Finally, we prove Lemma \ref{lem:KLD_bound} in Section \ref{subsec:KLD-bound}.

\subsection{KLD minimisation\label{subsec:app_KLD-minimisation}}

The augmented single trajectory space $\mathbb{\mathbb{U}}_{k'|k}\times T_{\left(k'\right)}$
can be written as the union of disjoint spaces $\mathbb{\mathbb{U}}_{k'|k}\times T_{\left(k'\right)}=\uplus_{u=0}^{n_{k'|k}}\left\{ u\right\} \times T_{\left(k'\right)}$.
Therefore, given a finite set $\widetilde{\mathbf{X}}_{k'}\subset\mathbb{\mathbb{U}}_{k'|k}\times T_{\left(k'\right)}$,
we can write $\widetilde{\mathbf{X}}_{k'}=\widetilde{\mathbf{Y}}_{k'}\uplus\widetilde{\mathbf{X}}_{k'}^{1}\uplus...\uplus\widetilde{\mathbf{X}}_{k'}^{n_{k'|k}}$,
where $\widetilde{\mathbf{Y}}_{k'}\subset\left\{ 0\right\} \times T_{\left(k'\right)}$
and $\widetilde{\mathbf{X}}_{k'}^{i}\subset\left\{ i\right\} \times T_{\left(k'\right)}$,
to obtain \cite[Eq. (3.53)]{Mahler_book14}
\begin{align}
 & \mathrm{D}\left(\widetilde{f}\left\Vert \widetilde{q}\right.\right)\nonumber \\
 & \,=\int\widetilde{f}\left(\widetilde{\mathbf{X}}_{k^{'}}\right)\log\frac{\widetilde{f}\left(\widetilde{\mathbf{X}}_{k^{'}}\right)}{\widetilde{q}\left(\widetilde{\mathbf{X}}_{k^{'}}\right)}\delta\widetilde{\mathbf{X}}_{k^{'}}\nonumber \\
 & \,=\int\int...\int\widetilde{f}\left(\widetilde{\mathbf{Y}}_{k'}\uplus\widetilde{\mathbf{X}}_{k'}^{1}\uplus...\uplus\widetilde{\mathbf{X}}_{k'}^{n_{k'|k}}\right)\nonumber \\
 & \,\times\log\frac{\widetilde{f}\left(\widetilde{\mathbf{Y}}_{k'}\uplus\widetilde{\mathbf{X}}_{k'}^{1}\uplus...\uplus\widetilde{\mathbf{X}}_{k'}^{n_{k'|k}}\right)}{\widetilde{q}\left(\widetilde{\mathbf{Y}}_{k'}\uplus\widetilde{\mathbf{X}}_{k'}^{1}\uplus...\uplus\widetilde{\mathbf{X}}_{k'}^{n_{k'|k}}\right)}\delta\widetilde{\mathbf{Y}}_{k'}\delta\widetilde{\mathbf{X}}_{k'}^{1}...\delta\widetilde{\mathbf{X}}_{k'}^{n_{k'|k}}\nonumber \\
 & =c-\int\widetilde{f}_{k'|k}^{p}\left(\widetilde{\mathbf{Y}}_{k'}\right)\log\widetilde{q}^{p}\left(\widetilde{\mathbf{Y}}_{k'}\right)\delta\widetilde{\mathbf{Y}}_{k'}\nonumber \\
 & \quad-\sum_{i=1}^{n_{k'|k}}\int\sum_{a\in\mathcal{A}_{k'|k}}w_{k'|k}^{a}\widetilde{f}_{k'|k}^{i,a^{i}}\left(\widetilde{\mathbf{X}}_{k'}^{i}\right)\log\widetilde{q}^{i,1}\left(\widetilde{\mathbf{X}}_{k'}^{i}\right)\delta\widetilde{\mathbf{X}}_{k'}^{i}
\end{align}
where $c$ is a constant that does not depend on $\widetilde{q}\left(\cdot\right)$.
Maximising with respect to $\widetilde{q}^{p}\left(\cdot\right)$,
$\widetilde{q}^{1,1}\left(\cdot\right)$, ...,$\widetilde{q}^{n_{k'|k},1}\left(\cdot\right)$,
we get
\begin{align}
\widetilde{q}^{p}\left(\widetilde{\mathbf{Y}}_{k'}\right) & =\widetilde{f}_{k'|k}^{p}\left(\widetilde{\mathbf{Y}}_{k'}\right)\\
\widetilde{q}^{i,1}\left(\widetilde{\mathbf{X}}_{k'}^{i}\right) & =\sum_{a\in\mathcal{A}_{k'|k}}w_{k'|k}^{a}\widetilde{f}_{k'|k}^{i,a^{i}}\left(\widetilde{\mathbf{X}}_{k'}^{i}\right).
\end{align}
Using (\ref{eq:Bernoulli_density_filter-aux}) and identifying terms
w.r.t. (\ref{eq:TPMB_approx}), we have that the existence probability
$r^{i}$ and single-trajectory density $p^{i}\left(\cdot\right)$
of $\widetilde{q}^{i,1}\left(\cdot\right)$ are
\begin{align}
r^{i} & =\sum_{a\in\mathcal{A}_{k'|k}}w_{k'|k}^{a}r_{k'|k}^{i,a^{i}}\label{eq:r_i_pmb_appendix}\\
p^{i}\left(X\right) & =\frac{\sum_{a\in\mathcal{A}_{k'|k}}w_{k'|k}^{a}r_{k'|k}^{i,a^{i}}p_{k'|k}^{i,a^{i}}\left(X\right)}{\sum_{a\in\mathcal{A}_{k'|k}}w_{k'|k}^{a}r_{k'|k}^{i,a^{i}}}.\label{eq:p_i_pmb_appendix}
\end{align}
By grouping similar local hypotheses, $r^{i}$ and $p^{i}\left(\cdot\right)$
become those given in Proposition \ref{prop:PMB_KLD_minimisation},
which finishes the proof.

\subsection{Matching the PHD\label{subsec:Matching-the-PHD}}

In this section, we show that, for KLD minisimation in the previous
section, it holds that the PHD of $\widetilde{f}\left(\cdot\right)$
is the same as the PHD of $\widetilde{q}\left(\cdot\right)$. In Sections
\ref{subsec:PHD-of_f} and \ref{subsec:PHD-of-q}, we calculate the
PHD of $\widetilde{f}\left(\cdot\right)$ and $\widetilde{q}\left(\cdot\right)$,
respectively. 

\subsubsection{PHD of $\widetilde{f}\left(\cdot\right)$\label{subsec:PHD-of_f}}

The PHD $D_{\widetilde{f}}\left(\cdot\right)$ of density $\widetilde{f}\left(\cdot\right)$
for sets of trajectories, which is given by (\ref{eq:PMBM_aux_var2}),
is \cite{Mahler_book14,Angel19_f}
\begin{align}
D_{\widetilde{f}}\left(\widetilde{X}\right) & =\int\widetilde{f}\left(\left\{ \widetilde{X}\right\} \cup\widetilde{\mathbf{X}}_{k'}\right)\delta\widetilde{\mathbf{X}}_{k'}\nonumber \\
 & =\int\widetilde{f}\left(\left\{ \widetilde{X}\right\} \cup\widetilde{\mathbf{Y}}_{k'}\uplus\widetilde{\mathbf{X}}_{k'}^{1}\uplus...\uplus\widetilde{\mathbf{X}}_{k'}^{n_{k'|k}}\right)\nonumber \\
 & \quad\delta\widetilde{\mathbf{Y}}_{k'}\delta\widetilde{\mathbf{X}}_{k'}^{1}...\delta\widetilde{\mathbf{X}}_{k'}^{n_{k'|k}}\label{eq:PHD_f_app}
\end{align}
where we have applied the decomposition of the set integral into disjoint
spaces, see Section \ref{subsec:app_KLD-minimisation}.

If $\widetilde{X}=\left(0,X\right)\in\left\{ 0\right\} \times T_{\left(k'\right)}$,
then the PHD is 
\begin{align}
D_{\widetilde{f}}\left(\widetilde{X}\right) & =\int\widetilde{f}_{k'|k}^{p}\left(\left\{ \widetilde{X}\right\} \cup\widetilde{\mathbf{Y}}_{k'}\right)\sum_{a\in\mathcal{A}_{k'|k}}w_{k'|k}^{a}\nonumber \\
 & \,\times\prod_{i=1}^{n_{k'|k}}\left[\widetilde{f}_{k'|k}^{i,a^{i}}\left(\widetilde{\mathbf{X}}_{k'}^{i}\right)\right]\delta\widetilde{\mathbf{Y}}_{k'}\delta\widetilde{\mathbf{X}}_{k'}^{1}...\delta\widetilde{\mathbf{X}}_{k'}^{n_{k'|k}}\nonumber \\
 & =\int\widetilde{f}_{k'|k}^{p}\left(\left\{ \widetilde{X}\right\} \cup\widetilde{\mathbf{Y}}_{k'}\right)\delta\widetilde{\mathbf{Y}}_{k'}\nonumber \\
 & =\lambda_{k'|k}\left(X\right)\label{eq:PHD_f_1}
\end{align}
where the last line follows directly as it corresponds to the PHD
of the PPP $\widetilde{f}_{k'|k}^{p}\left(\cdot\right)$.

If $\widetilde{X}=\left(u,X\right)\in\left\{ u\right\} \times T_{\left(k'\right)}$,
$u\in\left\{ 1,...,n_{k'|k}\right\} $, then the PHD is
\begin{align}
D_{\widetilde{f}}\left(\widetilde{X}\right) & =\int\widetilde{f}_{k'|k}^{p}\left(\widetilde{\mathbf{Y}}_{k'}\right)\sum_{a\in\mathcal{A}_{k'|k}}w_{k'|k}^{a}\widetilde{f}_{k'|k}^{u,a^{u}}\left(\left\{ \widetilde{X}\right\} \cup\widetilde{\mathbf{X}}_{k'}^{u}\right)\nonumber \\
 & \,\times\prod_{i=1,i\neq u}^{n_{k'|k}}\left[\widetilde{f}_{k'|k}^{i,a^{i}}\left(\widetilde{\mathbf{X}}_{k'}^{i}\right)\right]\delta\widetilde{\mathbf{Y}}_{k'}\delta\widetilde{\mathbf{X}}_{k'}^{1}...\delta\widetilde{\mathbf{X}}_{k'}^{n_{k'|k}}\nonumber \\
 & =\sum_{a\in\mathcal{A}_{k'|k}}w_{k'|k}^{a}\int\widetilde{f}_{k'|k}^{u,a^{u}}\left(\left\{ \widetilde{X}\right\} \cup\widetilde{\mathbf{X}}_{k'}^{u}\right)\delta\widetilde{\mathbf{X}}_{k'}^{u}\nonumber \\
 & =\sum_{a\in\mathcal{A}_{k'|k}}w_{k'|k}^{a}r_{k'|k}^{u,a^{u}}p_{k'|k}^{u,a^{u}}\left(X\right)\label{eq:PHD_f_2}
\end{align}

\subsubsection{PHD of $\widetilde{q}\left(\cdot\right)$\label{subsec:PHD-of-q}}

We calculate the PHD of $\widetilde{q}\left(\cdot\right)$, which
is given by (\ref{eq:TPMB_approx}) and Proposition \ref{prop:PMB_KLD_minimisation},
using (\ref{eq:PHD_f_app}). If $\widetilde{X}=\left(0,X\right)\in\left\{ 0\right\} \times T_{\left(k'\right)}$,
then the PHD is
\begin{align}
D_{\widetilde{q}}\left(\widetilde{X}\right) & =\int\int...\int\widetilde{q}^{p}\left(\left\{ \widetilde{X}\right\} \cup\widetilde{\mathbf{Y}}_{k'}\right)\nonumber \\
 & \:\times\prod_{i=1}^{n_{k'|k}}\left[\widetilde{q}^{i,1}\left(\widetilde{\mathbf{X}}_{k'}^{i}\right)\right]\delta\widetilde{\mathbf{Y}}_{k'}\delta\widetilde{\mathbf{X}}_{k'}^{1}...\delta\widetilde{\mathbf{X}}_{k'}^{n_{k'|k}}\nonumber \\
 & =\int\widetilde{q}^{p}\left(\left\{ \widetilde{X}\right\} \cup\widetilde{\mathbf{Y}}_{k'}\right)\delta\widetilde{\mathbf{Y}}_{k'}\nonumber \\
 & =\lambda_{k'|k}\left(X\right).\label{eq:PHD_q_1}
\end{align}
If $\widetilde{X}=\left(u,X\right)\in\left\{ u\right\} \times T_{\left(k'\right)}$,
$u\in\left\{ 1,...,n_{k'|k}\right\} $, then the PHD is
\begin{align}
D_{\widetilde{q}}\left(\widetilde{X}\right) & =\int\int...\int\widetilde{q}^{p}\left(\widetilde{\mathbf{Y}}_{k'}\right)\widetilde{q}^{u,1}\left(\left\{ \widetilde{X}\right\} \cup\widetilde{\mathbf{X}}_{k'}^{u}\right)\nonumber \\
 & \,\times\prod_{i=1,,i\neq u}^{n_{k'|k}}\left[\widetilde{q}^{i,1}\left(\widetilde{\mathbf{X}}_{k'}^{i}\right)\right]\delta\widetilde{\mathbf{Y}}_{k'}\delta\widetilde{\mathbf{X}}_{k'}^{1}...\delta\widetilde{\mathbf{X}}_{k'}^{n_{k'|k}}\nonumber \\
 & =\int\widetilde{q}^{u,1}\left(\left\{ \widetilde{X}\right\} \cup\widetilde{\mathbf{X}}_{k'}^{u}\right)\delta\widetilde{\mathbf{X}}_{k'}^{u}\nonumber \\
 & =\sum_{a\in\mathcal{A}_{k'|k}}w_{k'|k}^{a}r_{k'|k}^{u,a^{i}}p_{k'|k}^{u,a^{i}}\left(X\right).\label{eq:PHD_q_2}
\end{align}

We can see that the PHD of $\widetilde{q}\left(\cdot\right)$, which
is given by (\ref{eq:PHD_q_1}) and (\ref{eq:PHD_q_2}), coincides
with the PHD of $\widetilde{f}\left(\cdot\right)$, which is given
by (\ref{eq:PHD_f_1}) and (\ref{eq:PHD_f_2}).

\subsection{KLD bound\label{subsec:KLD-bound}}

In this section, we prove Lemma \ref{lem:KLD_bound}. We have
\begin{align}
\mathrm{D}\left(f_{k'|k}\left\Vert q\right.\right) & =\sum_{n=0}^{\infty}\frac{1}{n!}\int f_{k'|k}\left(\left\{ X_{1},...,X_{n}\right\} \right)\nonumber \\
 & \,\times\log\frac{f_{k'|k}\left(\left\{ X_{1},...,X_{n}\right\} \right)}{q\left(\left\{ X_{1},...,X_{n}\right\} \right)}dX_{1:n}\nonumber \\
 & =\sum_{n=0}^{\infty}\frac{1}{n!}\int\sum_{u_{1:n}}\widetilde{f}{}_{k'|k}\left(\left\{ \left(u_{1},X_{1}\right),...,\left(u_{n}X_{n}\right)\right\} \right)\nonumber \\
 & \,\times\log\frac{\sum_{u_{1:n}}\widetilde{f}{}_{k'|k}\left(\left\{ \left(u_{1},X_{1}\right),...,\left(u_{n}X_{n}\right)\right\} \right)}{\sum_{u_{1:n}}\widetilde{q}\left(\left\{ \left(u_{1},X_{1}\right),...,\left(u_{n}X_{n}\right)\right\} \right)}dX_{1:n}
\end{align}
where we have used (\ref{eq:integrate_out_auxiliary}). Applying the
log sum inequality \cite{Cover_book06} inside the integral, we obtain
\begin{align}
\mathrm{D}\left(f_{k'|k}\left\Vert q\right.\right) & \leq\sum_{n=0}^{\infty}\frac{1}{n!}\int\sum_{u_{1:n}}\widetilde{f}{}_{k'|k}\left(\left\{ \left(u_{1},X_{1}\right),...,\left(u_{n}X_{n}\right)\right\} \right)\nonumber \\
 & \,\times\log\frac{\widetilde{f}{}_{k'|k}\left(\left\{ \left(u_{1},X_{1}\right),...,\left(u_{n}X_{n}\right)\right\} \right)}{\widetilde{q}\left(\left\{ \left(u_{1},X_{1}\right),...,\left(u_{n}X_{n}\right)\right\} \right)}dX_{1:n}\nonumber \\
 & =\mathrm{D}\left(\widetilde{f}{}_{k'|k}\left\Vert \widetilde{q}\right.\right).
\end{align}
This completes the proof of Lemma \ref{lem:KLD_bound}.

\section{\label{sec:AppendixC}}

This appendix provides the expression of the mean and covariance matrix
of the density $p_{k|k}^{i}\left(\cdot\right)$ of updated Bernoulli
component $i$ after the GMTPMB updates. The case for the estimation
of the set of alive trajectories is given in Section \ref{subsec:App_Set-of-alive},
and the case of all trajectories in Section \ref{subsec:App_Set-of-all}. 

\subsection{Set of alive trajectories\label{subsec:App_Set-of-alive}}

The updated Bernoulli component $i$ of the resulting PMB density
from Lemma \ref{lem:Update_Gaussian_TPMB_alive} have $r_{k|k}^{i}$
given by (\ref{eq:existence_Prop2}), and mean and covariance matrix
\begin{align}
\overline{x}_{k|k}^{i} & =\sum_{a^{i}=1:r_{k|k}^{i,a^{i}}>0}^{h^{i}}\left[\frac{\overline{w}_{k|k}^{i,a^{i}}r_{k|k}^{i,a^{i}}}{r_{k|k}^{i}}\overline{u}_{k|k}^{i,a^{i}}\right]\label{eq:moment_matching_mean}\\
P_{k|k}^{i} & =\sum_{a^{i}=1:r_{k|k}^{i,a^{i}}>0}^{h^{i}}\left[\frac{\overline{w}_{k|k}^{i,a^{i}}r_{k|k}^{i,a^{i}}}{r_{k|k}^{i}}\left(W_{k|k}^{i,a^{i}}+\overline{u}_{k|k}^{i,a^{i}}\left(\overline{u}_{k|k}^{i,a^{i}}\right)^{T}\right)\right]\nonumber \\
 & \quad-\overline{x}_{k|k}^{i}\left(\overline{x}_{k|k}^{i}\right)^{T}\label{eq:moment_matching_cov}
\end{align}
where $\overline{w}_{k'|k}^{i,a^{i}}$ is given by (\ref{eq:weight_simplified_Prop2}),
which requires (\ref{eq:weight_global}) to relate to local hypotheses
weights. Local hypotheses with $r_{k|k}^{i,a^{i}}=0$ have no associated
$\overline{u}_{k|k}^{i,a^{i}}$ and $W_{k|k}^{i,a^{i}}$ so they are
not considered in the above sums.

\subsection{Set of all trajectories\label{subsec:App_Set-of-all}}

The updated Bernoulli component $i$ of the resulting PMB density
from Lemma \ref{lem:Update_Gaussian_TPMB_all} is of the form (\ref{eq:single_trajectory_Gaussian_all}).
For alive trajectories, $l=k$, we perform moment matching to obtain
$\overline{x}_{k|k}^{i}\left(k\right)$ and $P_{k|k}^{i}\left(k\right)$
in (\ref{eq:single_trajectory_Gaussian_all}), which yields
\begin{align}
\overline{x}_{k|k}^{i}\left(k\right) & =\sum_{a^{i}=1:r_{k|k}^{i,a^{i}}>0}^{h^{i}}\left[\frac{\overline{w}_{k|k}^{i,a^{i}}r_{k|k}^{i,a^{i}}\beta_{k|k}^{i,a^{i}}\left(k\right)}{\check{r}_{k|k}^{i}}\overline{u}_{k|k}^{i,a^{i}}\left(k\right)\right]\label{eq:mean_KLD_minimisation_all}\\
P_{k|k}^{i}\left(k\right) & =\sum_{a^{i}=1:r_{k|k}^{i,a^{i}}>0}^{h^{i}}\left[\frac{\overline{w}_{k|k}^{i,a^{i}}r_{k|k}^{i,a^{i}}\beta_{k|k}^{i,a^{i}}\left(k\right)}{\check{r}_{k|k}^{i}}\left(W_{k|k}^{i,a^{i}}\left(k\right)\right.\right.\nonumber \\
 & \left.\left.+\overline{u}_{k|k}^{i,a^{i}}\left(k\right)\left(\overline{u}_{k|k}^{i,a^{i}}\left(k\right)\right)^{T}\right)\right]-\overline{x}_{k|k}^{i}\left(k\right)\left(\overline{x}_{k|k}^{i}\left(k\right)\right)^{T}\label{eq:cov_KLD_minimisation_all}\\
\check{r}_{k|k}^{i} & =\sum_{a^{i}=1:r_{k|k}^{i,a^{i}}>0}^{h^{i}}\overline{w}_{k|k}^{i,a^{i}}r_{k|k}^{i,a^{i}}\beta_{k|k}^{i,a^{i}}\left(k\right).
\end{align}

\section{\label{sec:AppendixD}}

We proceed to determine the number of global hypotheses after the
first update for the filters based on Poisson birth model and multi-Bernoulli
birth model (labelled or not). We focus on the first update as the
number of global hypotheses is closed-form for all filters and we
can draw important insights into how the different filters deal with
hypotheses. We first consider updating a multi-Bernoulli RFS birth
with $n$ Bernoullis (with probability of existence $r\in\left(0,1\right)$)
with the MBM and MBM$_{01}$ ($\delta$-GLMB) filters. This result
is equivalent for filters based on targets and trajectories. Also,
the number of global hypotheses is not affected by labelling the MBM
and the MBM$_{01}$ \cite[Sec. IV]{Angel18_b}\cite[Sec. III.C]{Angel19_e}. 

Suppose we receive $m$ measurements. Then, for the MBM update (which
corresponds to the PMBM update with Poisson intensity equal to zero),
we obtain that the number of updated global hypotheses is
\begin{equation}
N_{A}^{\mathrm{MBM}}\left(m,n\right)=\sum_{p=0}^{\min\left(m,n\right)}p!\left(\begin{array}{c}
m\\
p
\end{array}\right)\left(\begin{array}{c}
n\\
p
\end{array}\right)\label{eq:Number_global_hypotheses_MBM}
\end{equation}
where $p$ represents the number of detected targets. The explanation
of (\ref{eq:Number_global_hypotheses_MBM}) is as follows. The number
of ways of selecting $p$ measurements from $m$ measurements is $\left(\begin{array}{c}
m\\
p
\end{array}\right)$. The number of ways of selecting $p$ targets from $n$ Bernoullis
is $\left(\begin{array}{c}
n\\
p
\end{array}\right)$. Finally, the number of ways of associating the detected measurements
with the detected targets is $p!$ and we must sum over all possible
$p$, which goes from 0 to $\min\left(m,n\right)$ to yield (\ref{eq:Number_global_hypotheses_MBM}). 

When we consider an MBM$_{01}$ ($\delta$-GLMB) update, the multi-Bernoulli
is converted into an MBM$_{01}$ ($\delta$-GLMB), in which targets
have deterministic existence rather than probabilistic. This step
results in an MBM$_{01}$ ($\delta$-GLMB) with $2^{n}$ components/global
hypotheses \cite[Sec. IV]{Angel18_b}\cite[Sec. IV.C.1]{Reuter14}.
In the update, each of these global hypotheses generates $N_{A}^{\mathrm{MBM}}\left(m,n_{a}\right)$
hypotheses, where $n_{a}$ is the number of alive targets in this
hypothesis. 

The number of MBM$_{01}$ global hypotheses with $n_{a}$ alive targets
is $\left(\begin{array}{c}
n\\
n_{a}
\end{array}\right).$ Therefore, the number of updated global hypothesis in MBM$_{01}$
($\delta$-GLMB) form is
\begin{equation}
N_{A}^{\mathrm{MBM}_{01}}\left(m,n\right)=\sum_{n_{a}=0}^{n}\left(\begin{array}{c}
n\\
n_{a}
\end{array}\right)N_{A}^{\mathrm{MBM}}\left(m,n_{a}\right).\label{eq:Number_global_hypotheses-MBM01}
\end{equation}
Table \ref{tab:Number-of-global_hyp_MBM_MBM01} shows the global hypotheses
for the PMBM, MBM and MBM$_{01}$ ($\delta$-GLMB) filters after the
first update. We have set $m=14$, as it is the average number of
measurements at the first time step given that all targets are detected
in the scenario in Section \ref{subsec:Linear_scenario}. The LMB
filter should first compute the number of $\delta$-GLMB updated components
(with pruning) and then apply the LMB approximation to the updated
$\delta$-GLMB. Both $\delta$-GLMB and LMB must prune a significant
number of global hypotheses for tractability. On the contrary, with
the Poisson birth model, which can handle an arbitrarily large number
of targets, the number of global hypotheses with a PMBM update is
1, i.e., it is already in PMB form. Therefore, PMBM and PMB (in targets
and trajectory spaces) do not lose any information in the first update
and require a lower computational time to keep the same information
in the posterior.

\begin{table}
\caption{\label{tab:Number-of-global_hyp_MBM_MBM01}Number of global hypotheses
after first update with an MB birth with $n$ Bernoullis and $m=14$,
and PPP birth. }

\begin{centering}
\begin{tabular}{c|c|c||c}
\hline 
\multicolumn{3}{c||}{MB birth} &
PPP birth\tabularnewline
\hline 
$n$ &
$\mathrm{MBM}$ &
$\mathrm{MBM}_{01}/\delta$-GLMB &
PMBM\tabularnewline
\hline 
4 &
33,909 &
46,328 &
\multirow{4}{*}{1}\tabularnewline
5 &
384,091 &
583,552 & \tabularnewline
6 &
4,010,455 &
6,882,352 & \tabularnewline
7 &
38,398,641 &
75,826,144 & \tabularnewline
\hline 
\end{tabular}
\par\end{centering}
\end{table}

\end{document}